\documentclass[preprint,11pt]{elsarticle}

\usepackage[a4paper, total={6in, 8in}]{geometry}
\usepackage{multicol}
\usepackage{amssymb}
\usepackage{amsmath}
\usepackage{amsmath}
\usepackage{subcaption}
\usepackage[mathscr]{euscript}
\usepackage{booktabs}
\usepackage{hyperref}

\usepackage[dvipsnames]{xcolor}

\journal{International Journal of Fatigue}

\begin{document}

\begin{frontmatter}

%% Title, authors and addresses

%% use the tnoteref command within \title for footnotes;
%% use the tnotetext command for theassociated footnote;
%% use the fnref command within \author or \affiliation for footnotes;
%% use the fntext command for theassociated footnote;
%% use the corref command within \author for corresponding author footnotes;
%% use the cortext command for theassociated footnote;
%% use the ead command for the email address,
%% and the form \ead[url] for the home page:
%% \title{Title\tnoteref{label1}}
%% \tnotetext[label1]{}
%% \author{Name\corref{cor1}\fnref{label2}}
%% \ead{email address}
%% \ead[url]{home page}
%% \fntext[label2]{}
%% \cortext[cor1]{}
%% \affiliation{organization={},
%%            addressline={}, 
%%            city={},
%%            postcode={}, 
%%            state={},
%%            country={}}
%% \fntext[label3]{}

\title{A Certifiable Machine Learning-Based Pipeline to Predict Fatigue Life of Aircraft Structures} %% Article title

%% use optional labels to link authors explicitly to addresses:
%% \author[label1,label2]{}
%% \affiliation[label1]{organization={},
%%             addressline={},
%%             city={},
%%             postcode={},
%%             state={},
%%             country={}}
%%
%% \affiliation[label2]{organization={},
%%             addressline={},
%%             city={},
%%             postcode={},
%%             state={},
%%             country={}}

\author[1]{Ángel Ladrón\corref{cor1}}  % ← corref va dentro del author
\ead{angel.lcordoba@upm.es}
\cortext[cor1]{Corresponding author.}
\author[1]{Miguel Sánchez-Domínguez}
\author[2]{Javier Rozalén}
\author[2]{Fernando R. Sánchez}
\author[1]{Javier de Vicente}
\author[3]{Lucas Lacasa}
\author[1,4]{Eusebio Valero}
\author[1,4]{Gonzalo Rubio}

%% Author affiliation
\affiliation[1]{organization={ETSIAE-UPM - School of Aeronautics, Universidad Politécnica de Madrid},%Department and Organization
            addressline={Plaza Cardenal Cisneros 3}, 
            postcode={E-28040}, 
            city={Madrid},
            country={Spain}}
            
\affiliation[2]{organization={Airbus Defence \& Space, Fatigue \& Damage Tolerance},%Department and Organization 
            city={Getafe},
            country={Spain}}
            
\affiliation[3]{organization={Institute for Cross-Disciplinary Physics and Complex Systems (IFISC), CSIC-UIB},%Department and Organization
            city={Palma de Mallorca},
            country={Spain}}
            
\affiliation[4]{organization={Center for Computational Simulation, Universidad Politécnica de Madrid, Campus de
Montegancedo, Boadilla del Monte},%Department and Organization
postcode={28660},
city={Madrid}, 
country={Spain}}

%% Abstract
\begin{abstract}
Fatigue life prediction is essential in both the design and operational phases of any aircraft, and in this sense safety in the aerospace industry requires early detection of fatigue cracks to prevent in-flight failures. Robust and precise fatigue life predictors are thus essential to ensure safety. 
Traditional engineering methods, while reliable, are time consuming and involve complex workflows, including steps such as conducting several Finite Element Method (FEM) simulations, deriving the expected loading spectrum, and applying cycle counting techniques like peak-valley or rainflow counting. %\cite{singh2010techniques}, 
%\cite{grover1966fatigue}. 
These steps often require collaboration between multiple teams and tools, added to the computational time and effort required to achieve fatigue life predictions. 
Machine learning (ML) offers a promising complement to traditional fatigue life estimation methods, enabling faster iterations and generalization, providing quick estimates that guide decisions alongside conventional simulations.
 In this paper, we present a ML-based pipeline that aims to estimate the fatigue life of different aircraft wing locations given the flight parameters of the different missions that the aircraft will be operating throughout its operational life. We validate the pipeline in a realistic use case of fatigue life estimation, yielding accurate predictions alongside a thorough statistical validation and uncertainty quantification. Our pipeline constitutes a complement to traditional methodologies by reducing the amount of costly simulations and, thereby, lowering the required computational and human resources.

\end{abstract}

%%Graphical abstract
% \begin{graphicalabstract}
%\includegraphics{grabs}
% \end{graphicalabstract}

%%Research highlights
% \begin{highlights}
% \item Research highlight 1
% \item Research highlight 2
% \end{highlights}

%% Keywords
\begin{keyword}
Fatigue Life \sep Machine Learning \sep Aerospace Safety \sep Aerospace Certification \sep Statistical validation
%% keywords here, in the form: keyword \sep keyword

%% PACS codes here, in the form: \PACS code \sep code

%% MSC codes here, in the form: \MSC code \sep code
%% or \MSC[2008] code \sep code (2000 is the default)

\end{keyword}

\end{frontmatter}

%% Add \usepackage{lineno} before \begin{document} and uncomment 
%% following line to enable line numbers
%% \linenumbers

%% main text
%%
%  \section*{List of Acronyms}
% \begin{tabular}{ll}
% \textbf{CFD} & Computational Fluid Dynamics \\
% \textbf{FEM} & Finite Element Method \\
% \textbf{G\&M} & Gusts and Maneuvers \\
% \textbf{GAG}  & Ground Air Ground \\
% \textbf{MAE} & Mean Absolute Error \\
% \textbf{ML}  & Machine Learning \\
% \textbf{MLP} & Multilayer Perceptron \\
% \textbf{MRE }& Mean Relative Error \\
% \textbf{PSE} & Principal Structural Element \\
% \end{tabular}
\section*{List of Acronyms}
\begin{tabular}{ll}
\textbf{AD} & Anderson–Darling \\
\textbf{CFD} & Computational Fluid Dynamics \\
\textbf{CI95} & 95\% Confidence Interval \\
\textbf{CLT} & Central Limit Theorem \\
\textbf{CV} & Coefficient of Variation \\
\textbf{EASA} & European Aviation Safety Agency \\
\textbf{ESDU} & Engineering Sciences Data Unit \\
\textbf{FEM} & Finite Element Method \\
\textbf{GAG} & Ground Air Ground \\
\textbf{G\&M} & Gusts and Maneuvers \\
\textbf{IQR} & Interquartile Range \\
\textbf{KS} & Kolmogorov–Smirnov \\
\textbf{MAE} & Mean Absolute Error \\
\textbf{ML} & Machine Learning \\
\textbf{MLP} & Multilayer Perceptron \\
\textbf{MRE} & Mean Relative Error \\
\textbf{PSE} & Principal Structural Element \\
\textbf{PSD} & Power Spectral Density \\
\end{tabular}

%% Use \section commands to start a section
\section{Introduction}
\label{sec:Introduction}

Damage estimation caused by fatigue \cite{grover1966fatigue} is a critical aspect of aeronautical engineering, as failure to predict it accurately can lead to catastrophic structural breakdowns and compromise both flight safety and operational efficiency. Likewise, fatigue damage assessment is essential for implementing predictive maintenance strategies that extend the life of aircraft and reduce unplanned downtime.
Standard fatigue prediction methods rely on a convoluted and resource-intensive process that combines service history of aircraft of similar structural design with aerodynamic and Finite Element Methods simulations and cumulative damage modeling to guide expert-based decision making (CS-25.571 in \cite{easaCS25}). The complexity of the whole process makes the development of tailored maintenance routines for new aircraft operators particularly expensive and time-consuming, involving multidisciplinary teams and a range of iterative procedures.

\medskip \noindent 
Interestingly, the advent of Machine Learning (ML) techniques and the rapid incorporation of these tools within the aeronautical engineering community (encompassing fields from aerodynamics \cite{linse1993identification, suresh2003lift, secco2017artificial}, stress prediction \cite{lacasa2025certification} or aeroacoustics \cite{ALGUACIL2021116285, ruettgers2020acoustic} to airflow optimization \cite{ramos2025transfer} --see \cite{LeClainche2023} for a recent review--)
provides a realistic opportunity to streamline and optimize the procedures for fatigue-related damage prediction, as these techniques potentially offer near-instantaneous inference of fatigue life, once properly trained.
Specifically in fatigue-related problems, ML has been recently used for load spectrum reconstruction \cite{graziani2024prediction, Valdes2017}, fatigue life prediction \cite{palczynski2022application, chen2023pinn}, and crack growth modeling \cite{Zhang2020, Younis2023}. However, most approaches treat these steps separately and rely on hybrid frameworks that combine ML with traditional tools, lacking end-to-end ML-enabled fatigue life prediction pipelines.\\
At the same time, there is an urge of incorporating fully certifiable ML-based procedures within the aeronautical industry. As a matter of fact, a key consideration for ML in safety-critical applications is the ability to certify the predictive model. Aeronautical authorities such as the European Aviation Safety Agency (EASA) have outlined the main challenges and goals of these industries for trustworthy ML adoption \cite{CoDANN, CoDANN2}, and recent research has started to propose practical methodologies for statistical validation of ML pipelines in industrial settings \cite{lacasa2025certification}, encompassing data preparation, model training and interpretability, and multiscale uncertainty quantification. 

\medskip 
\noindent 
This work proposes a fully integrated and certifiable ML-based pipeline capable of predicting fatigue life at various locations of an aircraft wing, based solely on flight parameters. The pipeline mostly uses a suite of different multilayer perceptrons (MLPs) to approximate different input-output relationships within the fatigue damage assessment.
An MLP is an archetypal method in supervised machine learning \cite{goodfellow2016deep} used to infer (regress or classify) complex input-output representations $x\to y$, where $x\in \mathbb{R}^m$ and $y$ is usually another vector for regression tasks or an element from a discrete set in classification tasks. Often visualized as an (artificial neural) network, an MLP consists in a (deep) feedforward neural network representing an overparameterized nonlinear function $\mathscr{F}(x;{\Omega})$, where $\Omega=\{w_k\}$ --the set of trainable parameters-- identify with the weights of the links between neurons (see  \cite{goodfellow2016deep} for details). The precise architecture of the MLP (number of neurons per layer, number of layers) as well as the precise optimization scheme to train the network are defined in a case-by-case basis.\\
Our approach enables end-to-end learning of the physical process --from stress estimation to load cycle characterization and fatigue damage inference-- reducing the need for intermediate, time consuming CFD or FEM simulations. At the same time, the pipeline enforces a statistically solid validation protocol which, overall, aims to pave the way for certifiable ML-driven fatigue assessment. Using real fatigue data from 38 wing locations of a tactical aircraft for different operational profiles, we show that the pipeline provides accurate fatigue life prediction along with a thorough characterization of the model's uncertainty.

\medskip \noindent
The rest of this paper is structured as follows: in Section \ref{sec:Methodology}, we present the full methodology: we define the physical problem under study, the dataset, the set of ML techniques and software tools, and the statistical techniques used for a thorough validation. Then, in Section \ref{sec:Results} we present the results, which we anticipate yield a successful ML-driven fatigue life prediction tool. 
Finally, conclusions and future directions are discussed in Section \ref{sec:Conclusions}.

%% Use \section commands to start a section
\section{Methodology}
\label{sec:Methodology}

In this section, we present the methodology adopted in this work. First, in subsections \ref{sec:met_prob} and \ref{sec:met_FL_ML}, we describe the physical problem being modeled using machine learning techniques and explain how it is decomposed into smaller, learnable subproblems. Then, in \autoref{sec:met_data_split}, we detail the available dataset and the data-splitting strategy used for training, which plays a crucial role in the performance of the models. Finally, in \autoref{sec:met_ML}, we add a few comments on the ML and statistical software libraries we have used.

\begin{figure}[h!]
    \centering
        \includegraphics[width=1.1\textwidth]{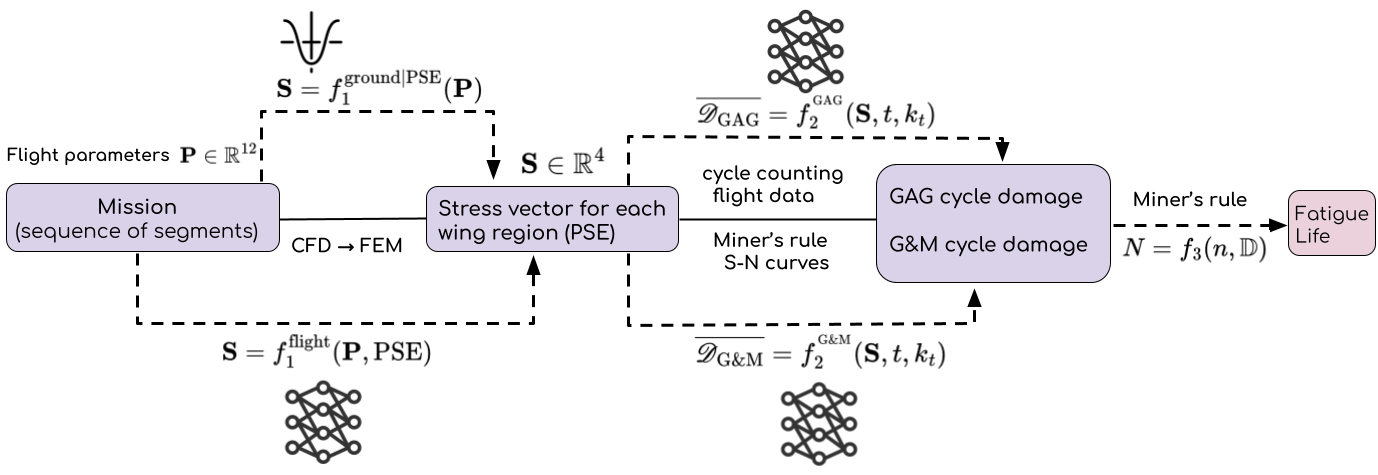}
    \caption{Illustration of the multi-step fatigue-related damage assessment (see the text for details). Our machine learning-based pipeline breaks down the whole process in three phases, where three function families $f_1=\{f_1^{\text{ground|PSE}}, f_1^{\text{flight}}\}$, $f_2=\{f_2^{\text{GAG}},f_2^{\text{G\&M}}\}, f_3$ are approximated with suitable statistical and deep learning models.}
    \label{fig:Traditional_meth_and_ML}
\end{figure}

\subsection{Breaking down the fatigue damage assessment}
\label{sec:met_prob}
The goal of this study is to develop a machine learning pipeline to predict fatigue life (expressed as the number of flights to failure) at multiple wing locations --known as principal structural elements (PSEs)--, when flying a specific mission (i.e., a flight characterized by specific flight parameters and operational objectives).  In the traditional life estimation method we aim to model, all PSEs are considered non-damage-tolerant, and consequently failure is defined as the number of flights until a detectable crack initiates. Here we adopt this very same definition.
In what follows we unfold the process and breakdown the whole supervised learning task in a sequence of interpretable subtasks, see Fig.~\ref{fig:Traditional_meth_and_ML} for an illustration of the different steps.

%\subsubsection{Traditional methodology}
%%% Explanation of traditional model

\begin{figure}[h!]
    \centering
    \includegraphics[width=.8\textwidth]{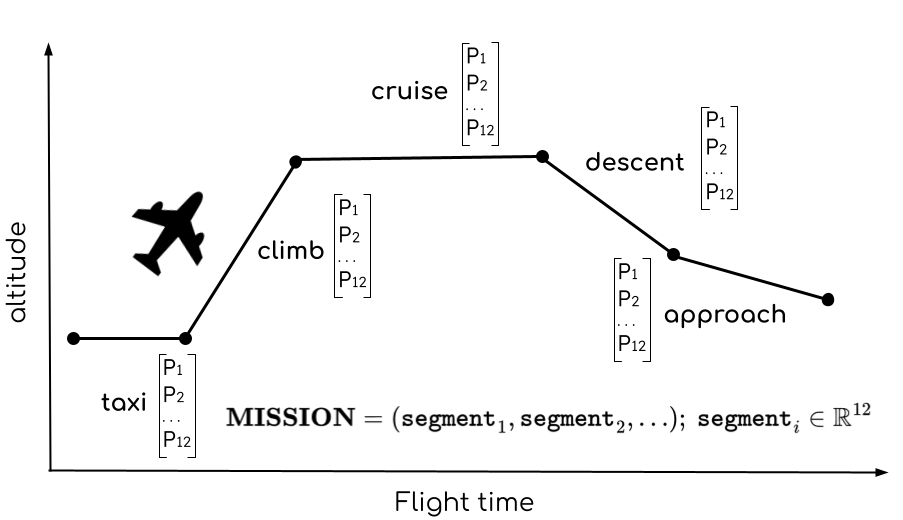}
    \caption{A mission flight plan, partitioned as a sequence of segments. Each segment is parametrized by a 12-dimensional feature vector $\bf P$ whose elements include variables such as the degree of extension of the flaps in that segment, the true airspeed during the segment, etc (see Table~\ref{tab:segment_variables} for details.)}
    \label{fig:mission}
\end{figure}

\medskip
\noindent {\bf Mission segmentation --} 
The assessment process starts by defining the mission plan (subparagraph 6b in CS-25.571, \cite{easaCS25}). This process consists in partitioning a mission into a structured sequence of flight segments, see Fig.~\ref{fig:mission}. Each of these segments characterize different specific phases of the flight, so that the categorical variable \textsc{segment} $\in$ \{\texttt{taxi}, \texttt{climb}, \texttt{cruise}, \texttt{descent}, \texttt{approach}\}. It is also useful to classify segments in the \texttt{ground phase} (\texttt{taxi}) or in the \texttt{flight phase} (every other segment except \texttt{taxi})\footnote{{While in rigor there exist ground segments other than taxi (e.g. \texttt{take-off}, \texttt{braking}, \texttt{rotation}, etc), these are not explicitly taken into account in the models as the wing PSEs don't show any significant damage induced by these type of segments.}}.
Accordingly, 
a mission flight plan is simply represented as a sequence of segments. At the same time, each segment is characterized by a vector of 12 flight parameters ${\bf P}\in \mathbb{R}^{12}$, that includes variables such as the degree of extension of the flaps in that segment, the true airspeed during the segment, the altitude at which the segment takes place, and so on (see Table~\ref{tab:segment_variables} for details). Accordingly, within each mission, any given flight is described in terms of a sequence of segment feature vectors $\{{\bf P}_i\}$ with class metadata (segment class and ground/air typification).

\medskip
\noindent {\bf From aerodynamic loads to PSE stresses --} Once the mission has been segmented and parameterized, Computational Fluid Dynamics (CFD) simulations are performed in each mission segment to obtain the corresponding aerodynamic loads. These loads are then propagated onto the structural level using Finite Element Method (FEM) simulations, so as to obtain the stresses at the different principal structural elements (PSEs) of interest (appendix 2 in CS-25.571, \cite{easaCS25}). The resulting values will be used for the stress spectrum generation. For each PSE, we have a particular four-dimensional stress vector ${\bf S} \in \mathbb{R}^4$, whose elements are ${\bf S}=(1g, \Delta \text{vman}, \Delta \text{vgust}, \Delta {\text{turn}})$, where $1g$ is the mean equilibrium stress at the studied PSE in steady flight (i.e. the structural response under nominal flight conditions), whereas the other three variables are incremental stresses at the studied PSE caused by a standard load factor of $1.5g$ due to vertical maneuvers, vertical gusts and turns. These latter variables capture the effects of atmospheric turbulence and pilot-induced maneuvers (see Table~\ref{tab:segment_stresses} for details). The incremental stresses values depend on the modeling. To account for gusts, Engineering Sciences Data Unit (ESDU) data or Power Spectral Density (PSD) models (CS-25.341 in \cite{easaCS25}) can be used, and for maneuvers, military regulations specify the requirements \cite{mil8861b}.
This step is computationally intensive and must be repeated for each mission segment individually, making it one of the most resource-consuming parts of the process. 
%The resulting values, known as $1g$ stresses, represent the structural response under nominal flight conditions. To account for additional loads induced by gusts and maneuvers, stress increments are introduced. 
%A rule of thumb is to estimate these increments using a fixed load factor of $1.5g$, capturing the effects of atmospheric turbulence and pilot-induced maneuvers.

%With the stress data obtained in the previous step and historical flight records from a similar aircraft, load cycles caused by gusts and maneuvers are constructed for each mission segment using techniques such as rainflow counting \cite{}. A load cycle corresponds to a complete stress variation between a maximum and a minimum value, and is represented as a vector comprising three elements: the maximum stress, the minimum stress, and the number of occurrences. In addition, the Ground-Air-Ground (GAG) cycle —which is critical for fatigue life but occur only once per flight— is defined based on the minimum and maximum stresses reached locally during the flight. The GAG cycle, along with the cycles induced by gusts and maneuvers are the two main sources of fatigue damage.

\medskip
\noindent {\bf From stresses to GAG and G\&M cycles --} With the stress data obtained in the previous step and historical flight records from a similar aircraft, two types of load cycles are derived for each mission: the so-called Ground-Air-Ground (GAG) cycle and the cycles induced by gusts and maneuvers (G\&M). The GAG cycle represents the dominant stress fluctuation experienced as the aircraft transitions from ground to flight segments. This cycle typically involves the largest stress amplitude of the flight (minimum to maximum stress) and occurs once per flight.
In contrast, gusts and maneuvers generate multiple smaller cycles within each flight segment. These are constructed using statistical load distributions and serve as an approximation of the fatigue loads that the aircraft will encounter across all the mission flights. Each resulting cycle is represented as a triplet: the maximum stress, the minimum stress, and the number of times that specific cycle occurs. Together, the cycles induced by gusts and maneuvers, along with the GAG cycle (derived from them) constitute the two main sources of fatigue damage—typically with the GAG cycle being the dominant contributor.

%Fourth, given the load cycles and material data (including the S-N curve and the stress concentration factor, $k_t$, chapters 4 and 7 of \cite{stephens2000metal}, ) the fatigue damage and life of the component can be assessed. The fatigue life, within the traditional methodology, can be expressed as:
\medskip
\noindent {\bf Final assessment of fatigue life --} 
Once the load cycles have been defined, the fatigue damage and corresponding life of the structural component can be estimated using material properties and local stress information. A key input in this step is the so-called S–N curve, which describes the relationship between the stress (typically the stress amplitude) and the number of cycles to failure for a given material and stress concentration factor, $k_t$. The factor $k_t$ quantifies how much the local stress is amplified due to defects or geometric features such as holes, fillets, or notches.
Based on these parameters, along with the load history, the mission definition and the principal structural element (PSE), the fatigue life, quantified as number-of-flights-to-failure $N$, can be formally expressed as:
\begin{equation}\label{eq:Life function}
     N = \mathscr{F} (\text{material}, k_t, \text{historical data}, \text{PSE}, \text{mission data}, \text{stress data}). 
\end{equation}
A linear damage accumulation model (Miner's rule \cite{bland1946discussion}) is finally adopted to estimate fatigue damage. The core idea is that the damage caused by each stress cycle is independent and additive, and thus the total damage is simply the sum of these contributions. Under this model, the damage $\mathbb{D}$ accumulated after $n$ flights per mission is simply given by:
\begin{equation}\label{eq:FatigueLife}
    \mathbb{D} = \frac{n}{N},
\end{equation}
such that by definition, failure occurs when the number of flights reaches the critical value $n^*=N$, i.e. at $\mathbb{D}=1$.
Although this model simplifies the effects of plasticity and the interactions between loads, it remains widely used in aerospace applications and serves as the reference for ground truth values in this study. 

%This linear damage accumulation model assumes that each load cycle contributes a fraction of damage and that total damage is simply the sum of these contributions. 
 
\subsection{Fatigue life prediction using Machine Learning}\label{sec:met_FL_ML}

As previously outlined, fatigue life estimation relies on a sequential process that demands substantial computational and human resources. It typically requires a combination of high-fidelity CFD and FEM simulations, followed by load cycle reconstruction and counting techniques. This workflow is not only time-consuming but also difficult to scale. In contrast, machine learning offers the potential for near-instantaneous inference and significant reductions in both computational cost and operational complexity, hence the opportunity.\\
The rationale starts from assuming that $\mathscr{F}(\cdot)$ in Eq.~\ref{eq:Life function} is a deterministic function, and aim to approximate it using a suite of ML models. Now, after preliminary inspection it was clear that modeling $\mathscr{F}$ with a single neural network led to stagnant loss curves and suboptimal learning. This, together with the fact that including every step of the process into a single neural network hinders interpretability, led us to reformulate the solution as a sequential pipeline composed of three supervised learning (regression) tasks, such that
%This structured approach not only improves learning performance by simplifying each task but also aligns with the physical understanding of the fatigue process, thereby facilitating validation and certification for real-world industrial scenarios.
%\begin{figure}[!h]
%    \centering
%    \includegraphics[width=0.99\linewidth]{Figures/Pipeline/ML_Pipeline.pdf}
%    \caption{ML pipeline \textbf{Se va a modificar ligeramente (f1, f2 y f3 están bien puestas)}}
%    \label{fig:Pipeline}
%\end{figure}https://es.overleaf.com/project/685a704419d1993478d2ebd8/detacher#
%\autoref{fig:Pipeline} presents an overview of the proposed machine learning pipeline, 
the global task (approximate $\mathscr{F}$ in Eq.~\ref{eq:Life function}) can be seen as a function composition, i.e. conceptually we have $\mathscr{F}= f_3 \circ f_2\circ f_1$, where $f_1$, $f_2$ and $f_3$ are nonlinear functions mapping different stages of the fatigue damage assessment pipeline described in Fig.~\ref{fig:Traditional_meth_and_ML}. As we shall see, each of these functions are approximated by different ML and statistical models.\\
As in the traditional approach, the ML pipeline begins by translating the mission into a structured sequence of flight segments, each described by 12 constant flight parameters. These parameters capture the operational characteristics of the aircraft during each segment (e.g., taxi, climb, cruise) and serve as input features for the first stage of the pipeline. For each \(\{\text{mission}, \text{PSE} \}\) pair, the pipeline proceeds in three phases:

\begin{itemize}
    \item \textbf{Phase I – Stress prediction}: The objective of this phase is to predict the stress vector \(\mathbf{S}\in \mathbb{R}^{4}\) at each principal structural element (PSE) and for every mission segment, using the corresponding flight parameters \(\mathbf{P} \in \mathbb{R}^{12}\) as input. Details on the components of \(\mathbf{P}\) and \(\mathbf{S}\) are provided in \ref{sec:PhaseI_database}.

Formally, this task would involve learning a function $f_1$ such that
\[
f_1: \mathbb{R}^{12} \rightarrow \mathbb{R}^4, \quad \mathbf{S} = f_1(\mathbf{P},\text{PSE})
\]
where the mapping is conditioned on the PSE under study. 
To account for the distinct physical regimes involved, we unfold $f_1$ into two separate models $f_1^{\text{ground|PSE}}$ and $f_1^{\text{flight}}$ that model $f_1$ for ground and flight segments, respectively. Ground segments, dominated by static loads (e.g., fuel weight), exhibit relatively simple stress responses and therefore we chose $f_1^{\text{ground|PSE}}$ to be a different second-order polynomial model for each PSE, with only one input variable (fuel weight) and a single output variable ($1g$). 
In contrast, flight segments involve dynamic aerodynamic loading, requiring more flexible functional shapes. Accordingly, $f_1^{\text{flight}}$ is approximated by a multilayer perceptron (MLP). After feature selection, we keep six out of the 12 features in $\bf P$ as input variables, and we add an encoding of the PSE as an additional independent variable, i.e. a single MLP was trained to generalize across all flight segments and wing locations. The output of this MLP is the full, four dimensional stress vector $\bf S$ (additional details on model architecture, input selection, and preprocessing for $f_1^{\text{flight}}$ are provided in \ref{appendix:PhaseI}).
    
    \item \textbf{Phase II – Damage prediction}: %The goal is to predict the accumulated fatigue damage due to GAG cycles and due to gusts and maneuvers after $n$ flights of a mission at a specific PSE. 
    The second phase of the pipeline estimates, from the stress data, the fatigue damage accumulated at each PSE over \(n\) flights of a given mission. Since the particularities of each PSE are already encoded in the different stresses, PSE is not anymore an input variable at the stage.
    Now, there are two distinct damage components:
$\mathbb{D}_\text{GAG}$, the total (accumulated) damage caused by the Ground–Air–Ground (GAG) cycle after $n$ flights, and $\mathbb{D}_\text{G\&M}$, the total (accumulated) damage induced by gusts and maneuver loads during flight segments after $n$ flights. These are the two primary contributors to fatigue degradation in metallic aircraft structures. Since they are distinct, we separate the analysis and consider two separate tasks:

\medskip \noindent 
(i) The first task is that of predicting the typical `per-flight' damage caused by the Ground–Air–Ground (GAG) cycle.Defining $\mathscr{D}_\text{GAG}$ as a latent random variable that characterises the GAG damage within a single flight, our first task is predicting the average value of this latent random variable, $\overline{\mathscr{D}_\text{GAG}}$, such that later we will use $ \mathbb{D}_\text{GAG} = \overline{\mathscr{D}_\text{GAG}}\cdot n$ to estimate the total GAG damage (i.e. $\mathbb{D}_{\text{GAG}}$ is a random variable defined as the sum of $n$ independent and identically distributed random variables $\mathscr{D}_{\text{GAG}}$).
The prediction of $\overline{\mathscr{D}_\text{GAG}}$ uses as input variables the average stress values during ground and flight segments \(\bar{\mathbf{S}}_{\text{ground}} \in \mathbb{R}\) and \(\bar{\mathbf{S}}_{\text{flight}} \in \mathbb{R}^4\) (computed from the Phase I stress predictions), along the time the aircraft spends in flight and on the ground during a single flight, \(t_{\text{ground}}\), \(t_{\text{flight}} \in \mathbb{R}\) and the stress concentration factor, \(k_t \in \mathbb{R}\) (see \ref{appendix:PhaseII} for details). In principle this is a per-flight quantity and thus independent of $n$. However, although $n$ is large (\( n \in [800, 13{,}000] \)) and the average damage is expected to asymptotically converge to the true per-flight value-as per the law of large numbers, see \ref{appendix:Statistical_analysis} for more details-we also add $n$ as an input variable, to account to possible deviations from the true mean due to finite samples effects. Mathematically

% While in principle this is a per-flight quantity and thus independent of $n$, it is only independent of $n$ in the limit of large number of flights (when the average asymptotically converges, as per the law of large numbers), so we also add $n$ as an input variable. Mathematically, 
\[
f_2^{\text{GAG}}: \mathbb{R}^{9} \rightarrow \mathbb{R}, \quad 
\overline{\mathscr{D}_\text{GAG}} = f_2^{\text{GAG}}(\bar{\mathbf{S}}_{\text{ground}}, \bar{\mathbf{S}}_{\text{flight}}, 
t_{\text{flight}}, t_{\text{ground}}, k_t,n).
\]
As previously stated, the total GAG-related damage after $n$ flights is $\mathbb{D}_\text{GAG}=\overline{\mathscr{D}_\text{GAG}}\cdot n$.

\medskip \noindent 
(ii) Likewise, the second task is that of predicting the the typical, per-flight damage caused by the Gusts and Maneuvers (G\&M) cycles $\overline{\mathscr{D}_\text{G\&M}}$. This prediction uses the same inputs as before, except for the ground-related variables. Mathematically, we have
\[
f_2^{\text{G\&M}}: \mathbb{R}^{7} \rightarrow \mathbb{R}, \quad 
\overline{\mathscr{D}_\text{G\&M}} = f_2^{\text{G\&M}}(\bar{\mathbf{S}}_{\text{flight}}, 
t_{\text{flight}}, k_t,n).
\]
The total G\&M-related damage after $n$ flights $\mathbb{D}_\text{G\&M}=\overline{\mathscr{D}_\text{G\&M}}\cdot n$.

We train two different MLPs to approximate $f_2^\text{GAG}$ and $f_2^\text{G\&M}$.
Note that the use of averaged stress values instead of segment-wise cycle estimation is motivated by the high variability of load spectra across flights and the lack of deterministic predictability of individual stress cycles. Since missions typically involve a large number of flights (\( n \in [800, 13{,}000] \)), this formulation is justified by the Central Limit Theorem and the assumption of independent (Miner's rule), identically distributed flights (see \ref{appendix:Statistical_analysis} for more details).
Both models were trained using log-transformed damage values to address the wide dynamic range in the output space, and we use a Mean Absolute Error (MAE) loss function to penalize poor performance on high-damage cases, which are the most critical in fatigue life estimation. Full details on the data preparation, feature engineering, model architecture and split strategy are provided in \ref{appendix:PhaseII}.

    \item \textbf{Phase III - Life prediction}: The final phase computes the fatigue life $N$ defined as the number-of-flights-to-failure for a given mission and PSE. The model takes as inputs the number of flights, $n$, and the two sources of accumulated fatigue damage  $\mathbb{D}_{\text{GAG}} , \mathbb{D}_\text{G\&M}$ predicted in Phase II, such that $N=f_3(\mathbb{D}_{\text{GAG}} ,\mathbb{D}_\text{G\&M})$. 
    Rather than approximating $f_3$ via an additional ML model, we instead apply the simpler Miner’s rule directly, as described in \autoref{sec:met_prob}. This linear damage accumulation model assumes failure occurs when total damage reaches unity. Accordingly, the total damage $\mathbb{D} = \mathbb{D}_{\text{GAG}} + \mathbb{D}_{\text{G\&M}}$ and thus fatigue life is computed as:  

\begin{equation}\label{eq:Fatigue life}
 N = f_3(n,\mathbb{D}_{\text{GAG}} , \mathbb{D}_\text{G\&M})= \frac{n}{\mathbb{D}_{\text{GAG}} + \mathbb{D}_\text{G\&M}}.
\end{equation}

%As stated in \autoref{sec:met_prob}, Miner's rule was adopted. Therefore from \autoref{eq:FatigueLife}, 
%\begin{equation}\label{eq:Fatigue life}
%    N = f_3(n, \mathbf{D}) = \frac{n}{D_{GAG} + D_{G\&M}},
%\end{equation}

\end{itemize}

\subsection{Dataset and split description}
\label{sec:met_data_split}
Our dataset contains information regarding 38 wing PSEs, labeled 1, 2, $\dots$ to 38, four different $k_t$ values, and seven different missions operated by a specific certification of the studied tactical aircraft, labeled A, B, $\dots$ to G. Each mission represents accumulated data over a large number of flights, typically ranging between \( n \in [800, 13000] \). 

As stated in \autoref{sec:met_prob}, fatigue damage assessment requires conducting different steps (\autoref{fig:Traditional_meth_and_ML}), such as running FEM simulations, load cycles construction and finally, computing damage (and life) with the S-N curves. Our database, contains the output of each one of these steps, which will serve as the ground truth values for our ML approach. 

As previously discussed, missions are divided into operational segments such as \texttt{taxi}, \texttt{climb}, \texttt{cruise}, \texttt{descent}, and \texttt{approach}, being each mission segment described in the dataset by 12 scalar flight parameters collected into a vector \(\mathbf{P} \in \mathbb{R}^{12}\). These flight parameters induce certain stresses at each PSE (FEM output), represented by 4-dimensional stress vectors \(\mathbf{S} \in \mathbb{R}^4\), which represent the local stress at the component for different loading conditions. Details about \textbf{P} and \textbf{S} provided in \ref{sec:PhaseI_database}.  We leverage this data to address phase I from \autoref{sec:met_prob}.\\
In addition, for all the missions, the load cycles induced by gusts and maneuvers during every segment, derived from statistical load distributions (\autoref{sec:met_prob}), are available at each PSE, along with the GAG cycles and the respective fatigue damages (for the four $k_t$ values) accumulated over $n$ flights of the respective mission. The ground truth of fatigue life is easily computed with these damages. 

\medskip
\noindent {\bf Data split --} Instead of performing a random split on all the $38 \times 7 \times 4 = 1064$  $\{\text{PSE}, \text{Mission}, k_{t}\}$ combinations- which could result in data leakage and in some PSEs being absent from the test set -the following split was performed.
For each PSE, a set of five out of the seven missions have been randomly selected for the training set (for the four $k_t$ values, treated as numerical variables), one mission for the validation set and another mission for the test set. The validation set is used for hyperparameter tuning while the test set is used to assess the overall performance of the pipeline. Details on how this split is applied in each phase are provided in \ref{sec:PhaseI_split} and \ref{sec:PhaseII_split}.

\subsection{Note on software}\label{sec:met_ML}
%To address each stage of the pipeline, we employed machine learning (ML) algorithms, primarily multilayer perceptrons (MLP). The 
ML models were developed using \texttt{pyLOM} \cite{pyLOM}, a high-performance computing-enabled framework for parallel reduced-order modeling implemented in PyTorch \cite{paszke2019pytorch}.
%This repository integrates various deep learning architectures for model order reduction---such as variational autoencoders---and for surrogate modeling, including Kolmogorov--Arnold Networks (KAN), graph neural networks (GNN), and multilayer perceptrons (MLP), all implemented in PyTorch \cite{paszke2019pytorch}.
In addition, we also used standard Python libraries, such as \texttt{scikit-learn} \cite{scikit-learn} for polynomial regression and data preprocessing, \texttt{pandas} \cite{mckinney-proc-scipy-2010} and \texttt{NumPy} \cite{harris2020array} for dataset manipulation, and \texttt{matplotlib} \cite{Hunter:2007} for visualization. %Finally, we also make use of the \texttt{NGValidation} library, a validation library co-developed between Airbus and UPM which implements the ideas described in \cite{lacasa2025certification} merging methods from deep learning, statistical data science, and mathematical optimization.\textcolor{red}{NO TENGO CLARO SI PONER LO DE NGVALIDATION, HAY QUE CONSULTARLO.}

%\subsection{Validation techniques and libraries}
%\label{sec:met_val}
%Since this pipeline targets an industrial and safety-critical application, it must be statistically validated to meet certification standards. Regulatory authorities require not only accurate predictions, but also reliable and robust models grounded in a physical understanding of the problem. It is therefore essential to characterize the errors and behavior of the models to determine where, when, and how they can be applied.

%To that end, we followed some of the guidelines covered in \cite{lacasa2025certification}, where the authors develop a pipeline to statistically validate surrogate models for industrial applications. Specifically, throughout the following section, we make an emphasis on studying the error of the models beyond the usual aggregated metrics. To do this, we study the model's error distribution conditioned to some variables of interest (such as the PSEs, the $k_t$ and the model's outputs). This will help us characterize the behavior of the models under different conditions (i.e. different missions, wing locations, etc.). 

%We also make use of the \texttt{NGValidation} library, a validation library co-developed between Airbus and UPM, which implements the ideas described in \cite{lacasa2025certification} using methods from deep learning, statistical data science, and mathematical optimization.

%% Use \section commands to start a section
\section{Results}
\label{sec:Results}
In this section, we start by presenting the results of every step of the pipeline $f_1,f_2,f_3$, along with a thorough error quantification of each of the predictions. The relative error used throughout this section is the absolute percentage error: 
\begin{equation}
\text{Relative Error (\%)} = \left| \frac{\hat{y} - y}{y} \right| \times 100, 
\end{equation}
where \( \hat{y} \) and \( y \) are the predicted and true values respectively. 
%depiected in Fig, to determine whether we managed to obtain accurate predictions and, consequently, if each $f_i$ from \autoref{fig:Pipeline} was properly modeled. 

\subsection{Phase I: Stress predictions }\label{sec:res_phaseI}
First, we will study the results of the stress vector \textbf{S} predictions of every mission segment and for every PSE $\in $ Test set. This phase approximates $f_1$ in Fig.~\ref{fig:Traditional_meth_and_ML}.

\begin{table}[h!]
\centering
\begin{tabular}{lccccc}
\toprule
 & Mean Relative Error (\%) & Std & Q1 & Q2 & Q3 \\
\midrule
$1g$  & 0.04 & 0.03 & 0.019 & 0.03 & 0.06 \\
\bottomrule
\end{tabular}
\caption{Relative Error statistics for $\textbf{S}$ predictions for ground segments, displaying the mean relative error (relative error averaged over all PSEs and ground segments), the standard deviation, and the first three quartiles.}
\label{tab:ground_mre_stats}
\end{table}

\medskip \noindent 
{\bf Ground segments: $f_1^{\text{ground|PSE}}$ --} As stated in \autoref{sec:met_prob}, the predictions of the stress vector for ground segments (taxi) were addressed with second order polynomial regression models $f_1^\text{ground|PSE}$. As detailed in \ref{appendix:PhaseI}, after an initial feature selection we decide to use just one of the 12 flight parameters --namely, fuel weight $FW \in \mathbb{R}$--. Likewise, for ground segments the only output variable is
the equilibrium stress $1g \in \mathbb{R}$. Accordingly, for each PSE, $1g = f_1^\text{ground|PSE}(FW)$, where $f_1^\text{ground|PSE}$  is a simple quadratic polynomial such as: 

\[ f_1^\text{ground|PSE}(FW) = \beta_0 + \beta_1FW + \beta_2FW^{2}, \] 
where $ \beta_0$, $ \beta_1$ and $ \beta_2$ are the polynomial coefficients, obtained by minimizing the residual sum of squares between predictions and ground truth values (ordinary least squares) \cite{scikit-learn}.

\begin{figure}
    \centering
    \includegraphics[width=0.7\linewidth]{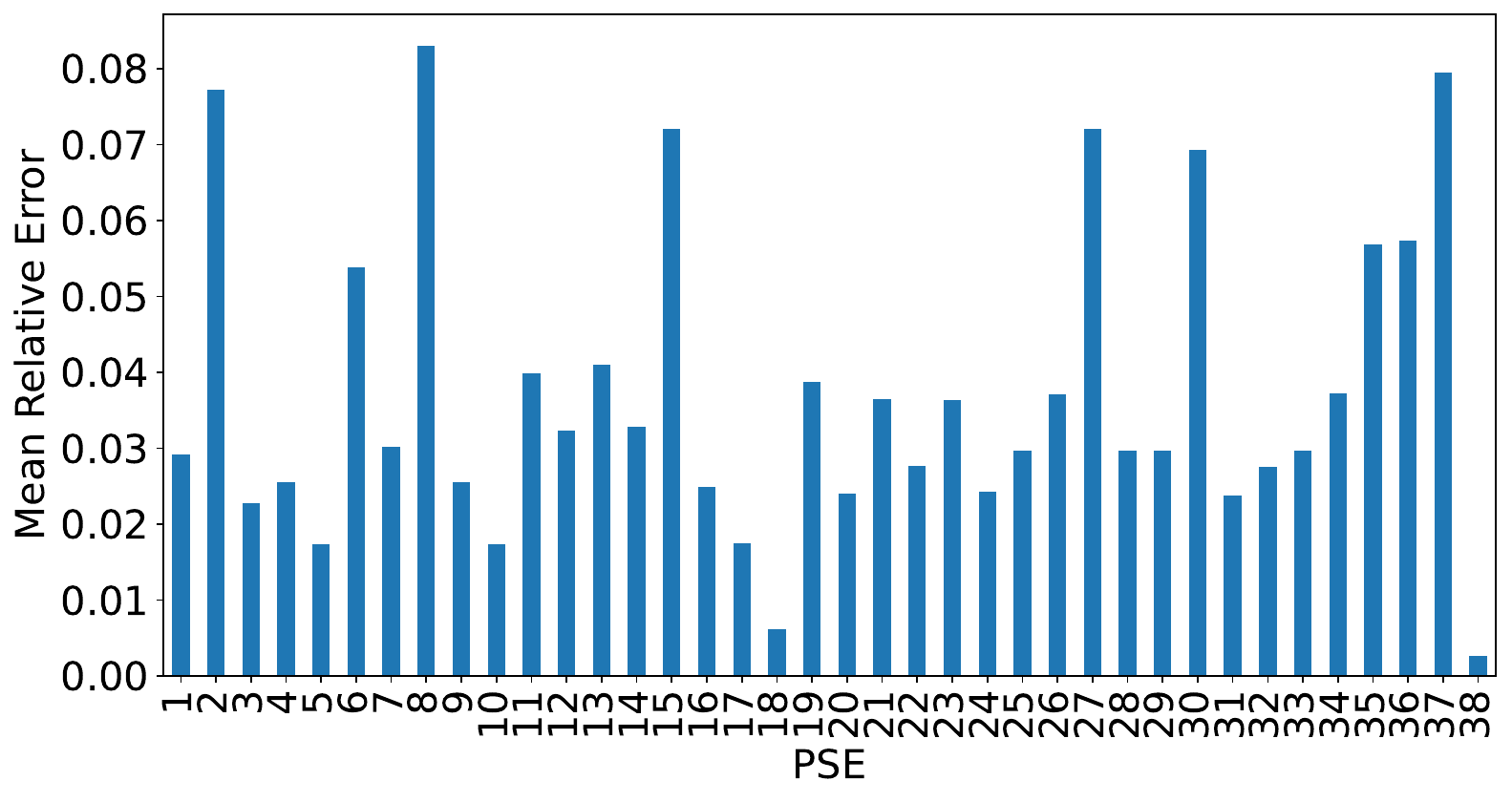}
    \caption{Mean relative error (in $\%$) for the predicted $1g$ for ground segments, grouped by PSEs.}
    \label{fig:ground_1g_PSEs}
\end{figure}

First, in Table~\ref{tab:ground_mre_stats} we display the statistics of the relative error for the $1g$ stress prediction in the test set. The mean relative error (MRE) --i.e. the relative error averaged over all ground segments and PSEs-- is substantially below $1\%$, indicating a very good performance. To gain insight on the dependency of the prediction error on the PSEs and missions, we depart from pointwise statistics and in
Fig.~\ref{fig:ground_1g_PSEs} we plot the Mean Relative Error (MRE)  of the $1g$ stress predictions conditioned on (i.e. grouped by) PSEs. As observed, although there is a slight variability in performance between PSEs, for all PSEs the MRE systematically remains below 0.1\%.  Similarly, \autoref{tab:ground_mre_mission} shows the performance (again in terms of MRE) but now grouped by missions instead of PSEs. As observed, all the missions exhibit excellent performance and again, MRE systematically remains below 0.1\%.  Overall, these results indicate that the dependence of the $1g$ stresses on flight parameters for ground segments is very simple and is well captured by a second order polynomial.
%suggest that the implicit input-output relationship for ground segments appears to be mild nonlinear, since a second order polynomial regressor managed to model it with very low errors. 

\begin{table}[h!]
\centering
\begin{tabular}{lccccccc}
\toprule
 & A & B & C & D & E & F & G \\
\midrule
$1g$  & 0.05 & 0.05 & 0.03 & 0.02 & 0.04 & 0.04 & 0.05 \\
\bottomrule
\end{tabular}
\caption{MRE for the predictions of each element of the stress vector $\textbf{S}$ over ground segments, grouped per mission class.}
\label{tab:ground_mre_mission}
\end{table}

\begin{figure}[h!]
    \centering
    \includegraphics[width=0.6\linewidth]{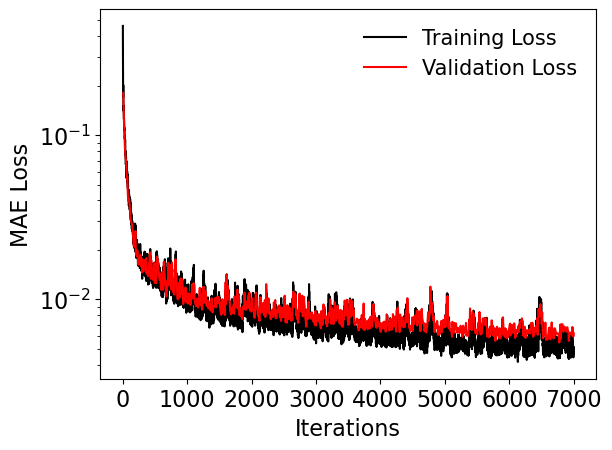}
    \caption{Learning curves (training and validation loss) for the MLP that approximates $f_1^{\text{flight}}$.} 

    \label{fig:unitarios_learning curve}
\end{figure}

\begin{figure}[h!]
    \centering

    % Primera fila
    \begin{subfigure}{0.49\textwidth}
        \centering
        \includegraphics[width=\textwidth]{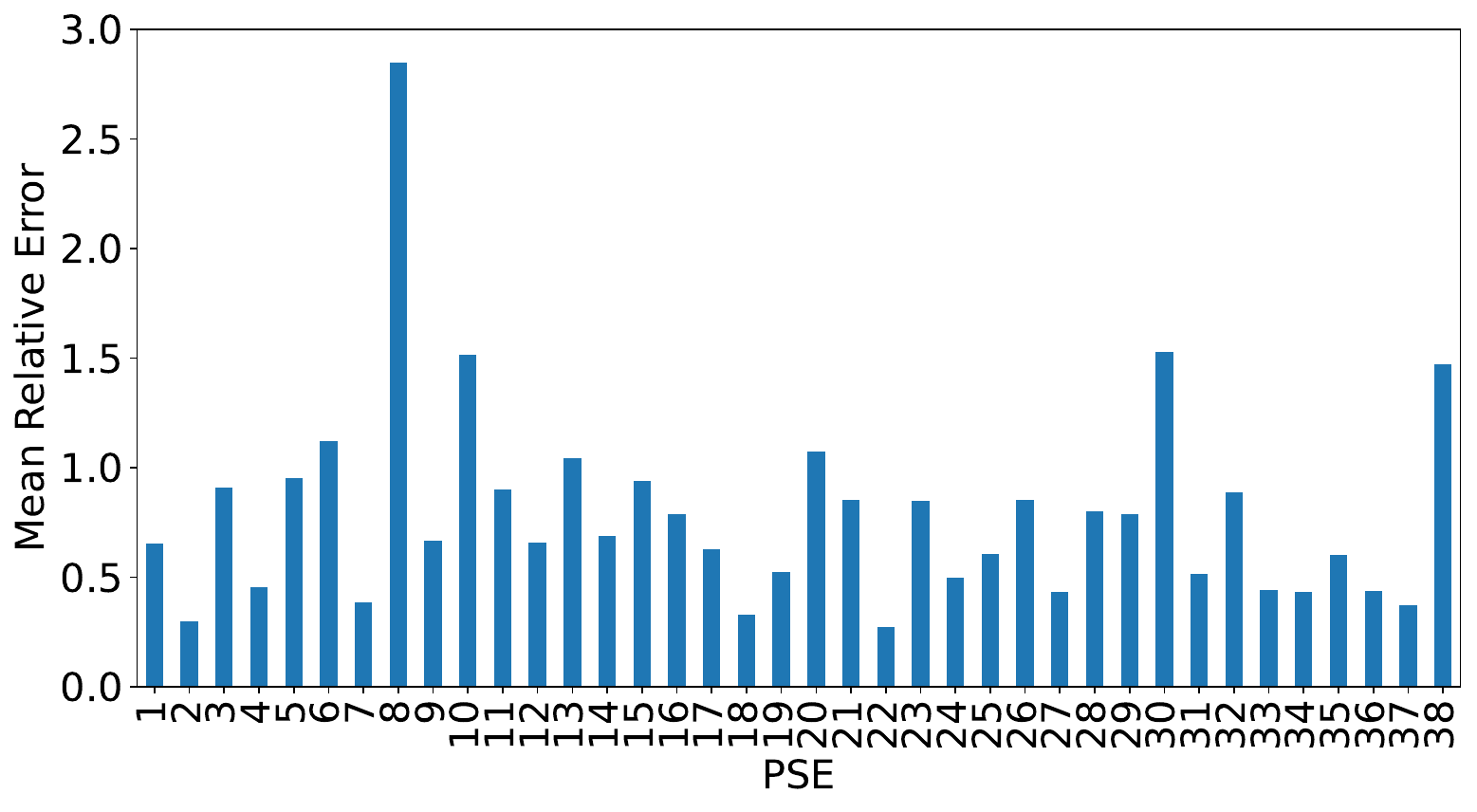}
        \caption{Mean relative error for $1g$ predictions for flight segments grouped by PSEs}
        \label{subfig:flight_1g_PSEs}
    \end{subfigure}
    \hfill
    \begin{subfigure}{0.49\textwidth}
        \centering
        \includegraphics[width=\textwidth]{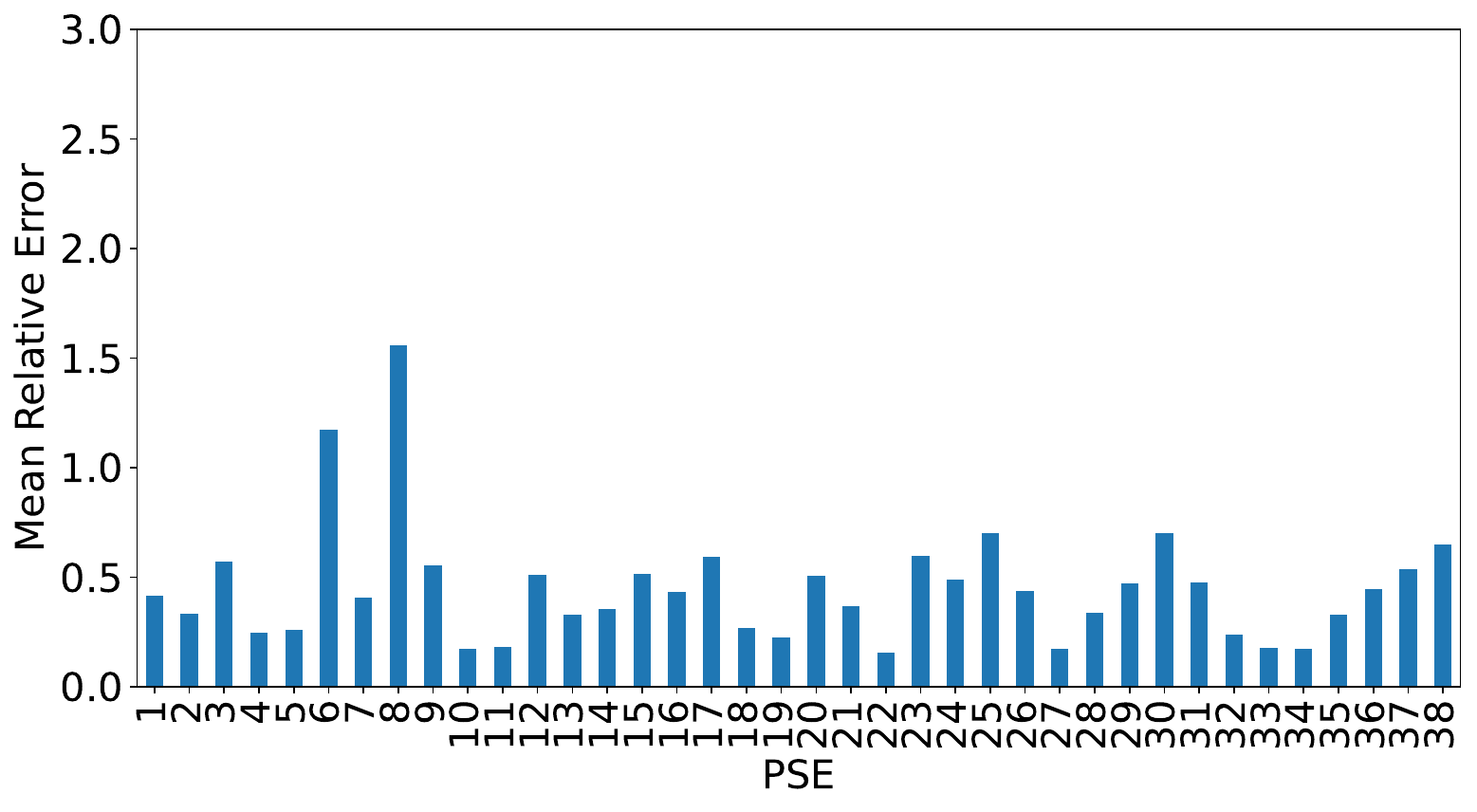}
        \caption{Mean relative error for $\Delta\text{vman}$ predictions for flight segments grouped by PSEs}
        \label{subfig:flight_Vman_PSEs}
    \end{subfigure}

    \vspace{0.5em} % Espacio vertical entre filas

    % Segunda fila
    \begin{subfigure}{0.49\textwidth}
        \centering
        \includegraphics[width=\textwidth]{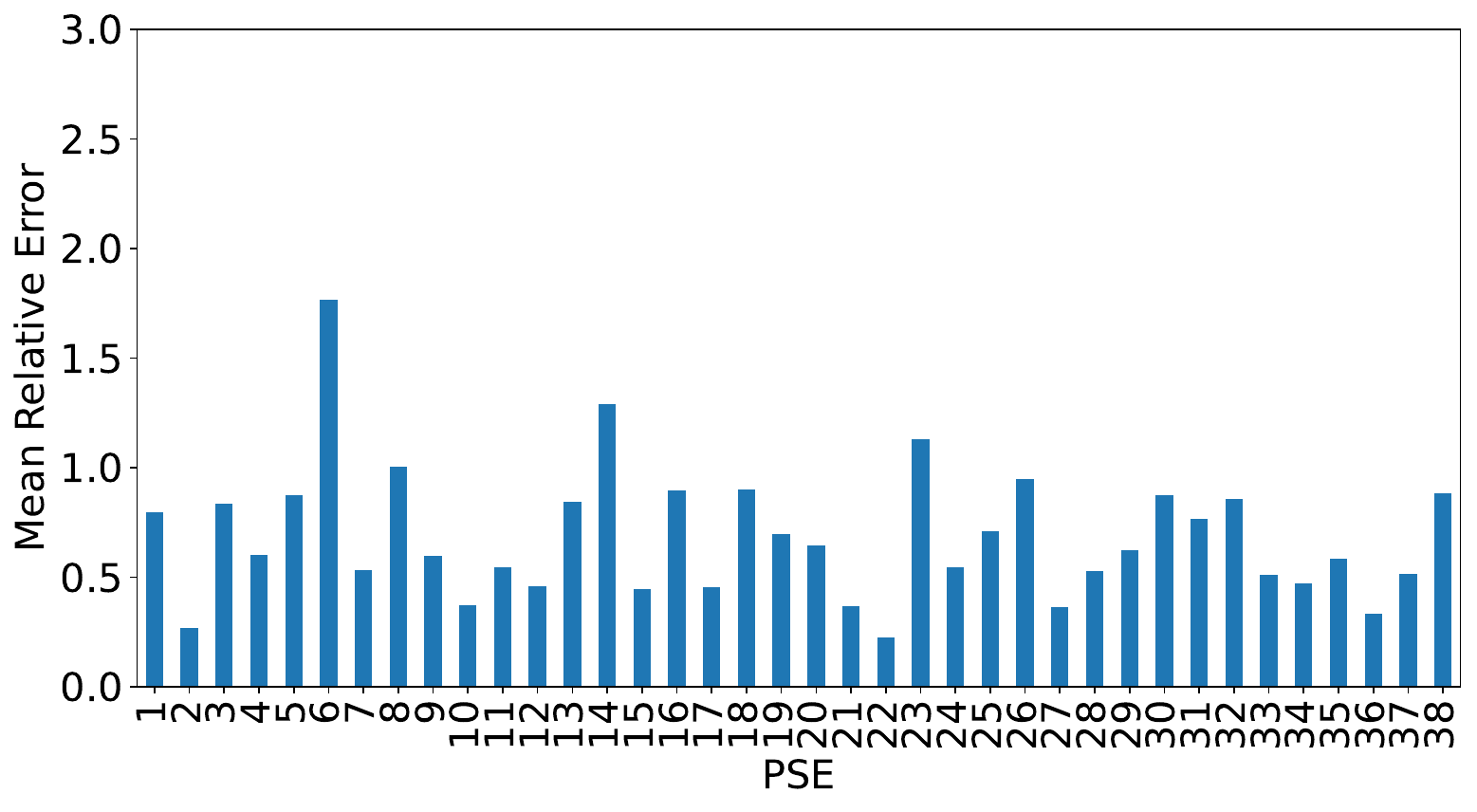}
        \caption{Mean relative error for $\Delta\text{vgust}$ predictions for flight segments grouped by PSEs}
        \label{subfig:flight_Vgust_PSEs}
    \end{subfigure}
    \hfill
    \begin{subfigure}{0.49\textwidth}
        \centering
        \includegraphics[width=\textwidth]{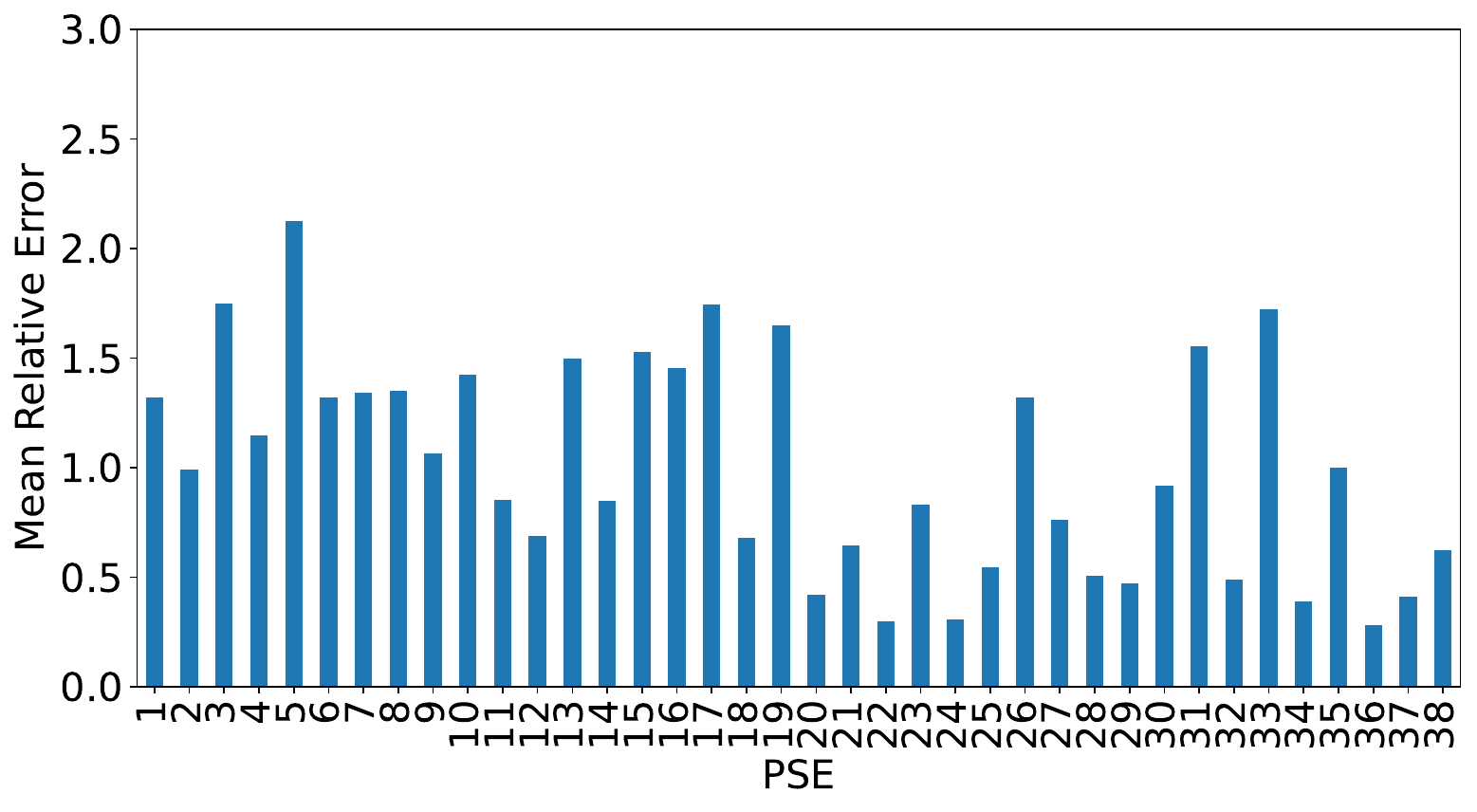}
        \caption{Mean relative error for $\Delta\text{turn}$ predictions for flight segments grouped by PSEs}
        \label{subfig:flight_Turn_PSEs}
    \end{subfigure}
    
    % Caption general
    \caption{Mean relative error grouped by PSEs for the all the predicted stresses}
    \label{fig:unitarios_PSEs}
\end{figure}

\begin{figure}[h!]
    \centering
    \includegraphics[width=0.99\linewidth]{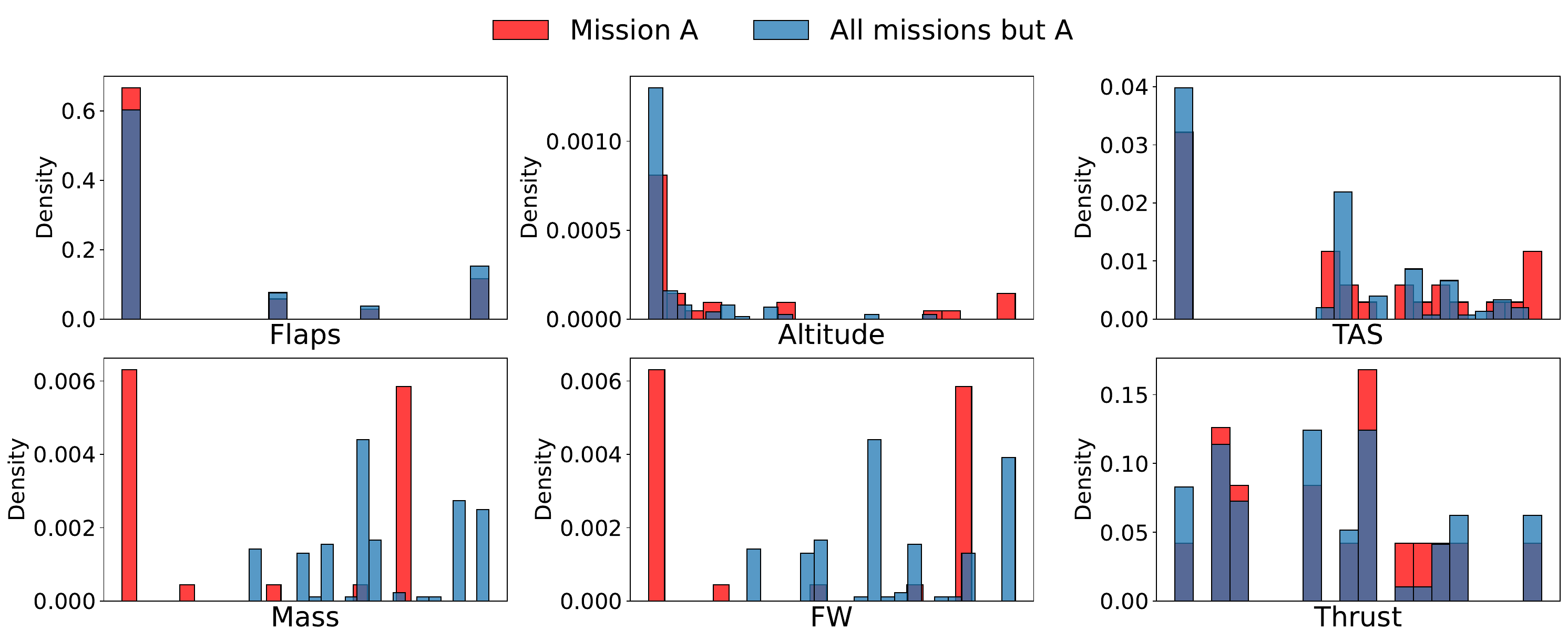}
    \caption{Histograms of input variables for mission A vs all the other missions. Specific support ranges have been hidden for confidentiality reasons.}
    \label{fig:params_hist}
\end{figure}

\medskip
\noindent 
{\bf Flight segments: $f_1^{\text{flight}}$ --}
 In \autoref{fig:unitarios_learning curve} we display the learning curves (MAE training and validation losses as a function of the number of iterations of the optimization scheme). As observed, both curves remain close to each other and after a sharp decrease they stabilize in a low value, hence showing a good learning behavior with no appreciable underfitting or overfitting. We then depict similar performance metrics for $f_1^{\text{ground|PSE}}$: relative error statistics for all four elements of $\bf S$ are depicted in Table~\ref{tab:mre_stats_globales}, whereas the MRE conditioned on PSEs and on mission classes are depicted in Fig.~\ref{fig:unitarios_PSEs} and Table~\ref{tab:mre_por_mision}, respectively. Although turn maneuvers show a slightly poorer performance, there is an accurate performance for almost all the PSEs and predicted stresses, and we can conclude that $f_1^{\text{flight}}$ is well approximated by the MLP model. Incidentally, we observe slightly worse performance when predicting mission A data. In hindsight, this deviation in performance can be easily explained by visualizing some of the flight parameter histograms for mission A, as compared to the rest of missions. This data is depicted in Fig.~\ref{fig:params_hist}, where we observe that for some of the flight parameters (altitude and some mass-related parameters), mission A's distributions deviate from the bulk for some segments, i.e. they are located in a poorly sampled region of the input space, a well-known reason for poorer generalization \cite{lacasa2025certification}. 

 \begin{table}[h!]
\centering
\begin{tabular}{lccccc}
\toprule
 & MRE (\%) & Std & Q1 & Q2 & Q3 \\
\midrule
$1g$              & 0.86 & 0.89 & 0.29 & 0.63 & 1.06 \\
$\Delta\text{vman}$   & 0.48 & 0.50 & 0.14 & 0.33 & 0.68 \\
$\Delta\text{vgust}$  & 0.73 & 0.66 & 0.25 & 0.56 & 1.03 \\
$\Delta\text{turn}$            & 1.17 & 1.05 & 0.40 & 0.87 & 1.60 \\
\bottomrule
\end{tabular}
\caption{Relative error statistics for the predictions of all four elements of the stress vector $\textbf{S}$ over flight segments: mean relative error (MRE, \%, averaging over all PSEs and flights segments), standard deviation, and the first three quartiles Q1 (percentile 25), Q2 (median) and Q3 (percentile 75).}
\label{tab:mre_stats_globales}
\end{table}

%In \autoref{fig:unitarios_SSEs} the mean relative error for each stress prediction, grouped by SSEs (1-38) is depicted. 

%Besides, \autoref{tab:mre_por_mision} and \autoref{tab:mre_stats_globales} present the same results, but grouped by missions (A-G) and the standard deviation and some percentiles of the relative errors of the predictions respectively. These results suggest that the implicit relationship, $\textbf{S} = f_1(\textbf{P})$, is correctly modeled. Once again, turn maneuvers show a poorer performance and there is a notably difference in performance for mission A. 

\begin{table}[h!]
\centering
\begin{tabular}{lccccccc}
\toprule
Mission & A & B & C & D & E & F & G \\
\midrule
$1g$              & 1.12 & 0.60 & 0.79 & 0.66 & 0.65 & 0.65 & 0.66 \\
$\Delta\text{vman}$   & 0.68 & 0.37 & 0.36 & 0.42 & 0.41 & 0.42 & 0.30 \\
$\Delta\text{vgust}$  & 0.90 & 0.52 & 0.68 & 0.70 & 0.53 & 0.69 & 0.57 \\
$\Delta\text{turn}$              & 1.29 & 0.56 & 1.66 & 0.97 & 0.48 & 0.64 & 0.75 \\
\bottomrule
\end{tabular}
\caption{MRE per mission for $\textbf{S}$ predictions over flight segments}
\label{tab:mre_por_mision}
\end{table}

Modeling this first phase of the pipeline with $\{f_1^{\text{ground|PSE}},f_1^{\text{flight}}\}$ has demonstrated good accuracy. The output of these models will serve to build the input for the following steps of the pipeline.
Observe that the small errors in the prediction of stress vectors will propagate into phase II prediction.
%Even though there are small errors introduce some noise in the next steps of the pipeline, addressing the second phase $f_2$, in which these predictions are needed to build the input, is feasible.

\subsection{Phase II: GAG and G\&M Damage prediction}\label{sec:res_phaseII}

\begin{figure}[h]
    \centering
    \begin{subfigure}{0.48\textwidth}
        \centering
        \includegraphics[width=\textwidth]{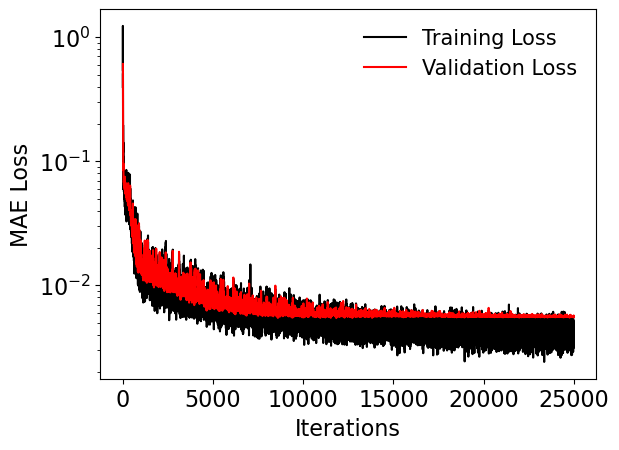}
        \caption{GAG model learning curves}
        \label{subfig:curves_GAG}
    \end{subfigure}
    % \hfill    
    \begin{subfigure}{0.48\textwidth}
        \centering
        \includegraphics[width=\textwidth]{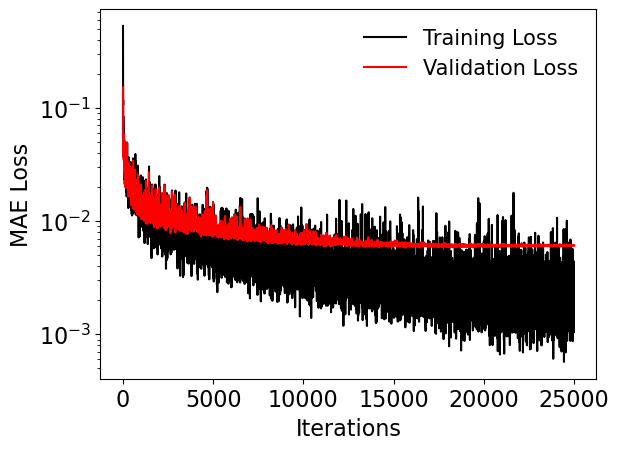}
        \caption{Gusts and maneuvers model learning curves}
        \label{subfig:GandM_learning_curves}
    \end{subfigure}
    \caption{Learning curves for both damage models approximating (a) $f_2^{\text{GAG}}$ and (b) $f_2^{\text{G\&M}}$.}
    \label{fig:damage_curves}
\end{figure}

In this phase, both sources of fatigue damage are predicted separately with two different MLP models, that approximate $f_2^{\text{GAG}}$ and $f_2^{\text{G\&M}}$ respectively. In Fig.~\ref{fig:damage_curves} the learning curves of the models are depicted. They were trained for 5000 epochs (note that working with mini-batches results in many iterations per epoch). As observed, during the training process, the validation loss remains close to the training loss (MAE) for both models. The gusts and maneuvers damage model shows a more noisy learning process than the GAG damage model, but still manages to learn. Finally, the validation sets remain flat, so extending the training for more epochs might lead to a high-variance scenario, where the model would learn spurious noise.

\medskip \noindent In \autoref{tab:mre_GAG} and \autoref{tab:mre_G&M} we show a summary of the relative error (\%) statistics of the damage predictions $\mathbb{D}_{\text{GAG}}$ and $\mathbb{D}_{\text{G\&M}}$, for all the points in the test set. As we can see, the mean relative error is moderately large and substantially larger that the median, what suggests a skewed, fat-tailed error distribution. Notice, indeed, that the third quartile is not symmetric with respect to the first quartile, and that the max error is disproportionally large. This suggests that pointwise error statistics are being deformed --and the average performance apparently decreased-- due to the presence of outliers. So it is important to ascertain where these outliers are located. Since the models predict fatigue damage --which can lead to instantaneous breakage--, greater damages represent more critical situations in both cases, as fatigue failure will be achieved earlier, resulting in shorter fatigue lives. 
Therefore, it is in the high-damage region that the MLP models must be more accurate, and therefore we need to certify that outliers are never present in this region.

%Note that these results combine the good performance in the region of interest with the poorer performance in the low fatigue damage region, which lowers the overall performance statistics of the model.

\begin{figure}[h!]
\centering
\begin{minipage}{0.49\textwidth}
    \centering
    \captionsetup{type=table} 
    \begin{tabular}{cc}
         \toprule
            Relative error stats  & Values \\
            \midrule
            Mean                          & 8.12 \\
            Standard deviation            & 12.10 \\
            First quartile (25\%)         & 1.21 \\
            Median (50\%)                 & 3.67 \\
            Third quartile (75\%)         & 9.40  \\
            Min error                     & 0.08 \\
            Max error                     & 80.95 \\
        \bottomrule
    \end{tabular}
        \caption{Relative error statistics for the prediction of $\mathbb{D}_{\text{GAG}}$.}
    \label{tab:mre_GAG}
\end{minipage}
\hfill
\begin{minipage}{0.49\textwidth}
    \centering
    \captionsetup{type=table} 
    \begin{tabular}{cc}
        \toprule
        Relative error stats & Values \\
        \midrule
        Mean                           & 9.49 \\
        Standard deviation             & 21.80 \\
        First quartile (25\%)          & 2.47 \\
        Median (50\%)                  & 4.94 \\
        Third quartile (75\%)          & 8.40 \\
        Min error                      & 0.03 \\
        Max error                      & 202.29 \\
        \bottomrule
    \end{tabular}
        \caption{Relative error statistics for the prediction of $\mathbb{D}_{\text{G\&M}}$.}
    \label{tab:mre_G&M}
\end{minipage}
\end{figure}

\begin{figure}[h!]
    \centering
    \includegraphics[width=0.8\linewidth]{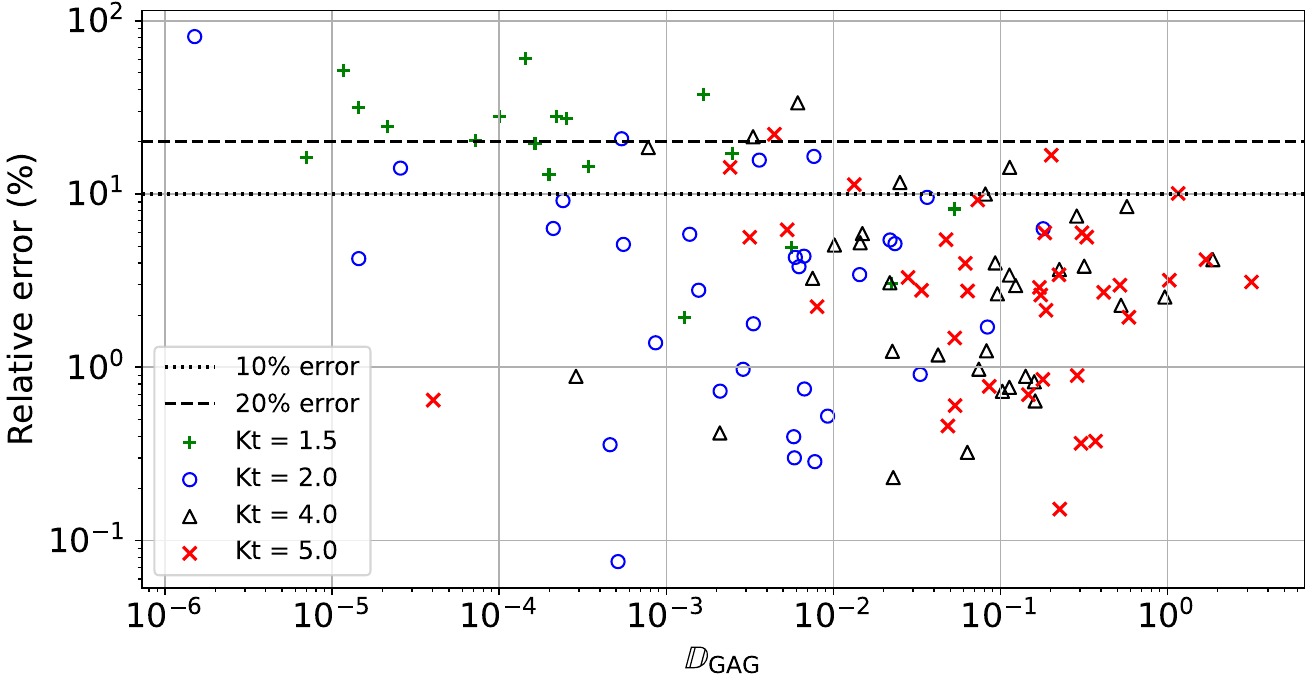}
    \caption{Scatter plot of the relative error of $\mathbb{D}_{\text{GAG}}$ as a function of its ground truth value, for every point in the test set.}
    \label{fig:mre_GAG}
    % TB: COMO DICES QUE HAY CORRELACION NEGATIVA ENTRE RELATIVE ERROR Y DAMAGE, XQ NO LA SACAS? MAS IMPORTANTE QUE EL COEFICIENTE DE CORRELACION SERA EL P-VALOR DEL TEST DE CORRELACION. USA MEJOR CORRELACION DE SPEARMAN PUES EL PLOT ES LOG-LOG
\end{figure}

\begin{figure}[h!]
    \centering
    \includegraphics[width=0.8\linewidth]{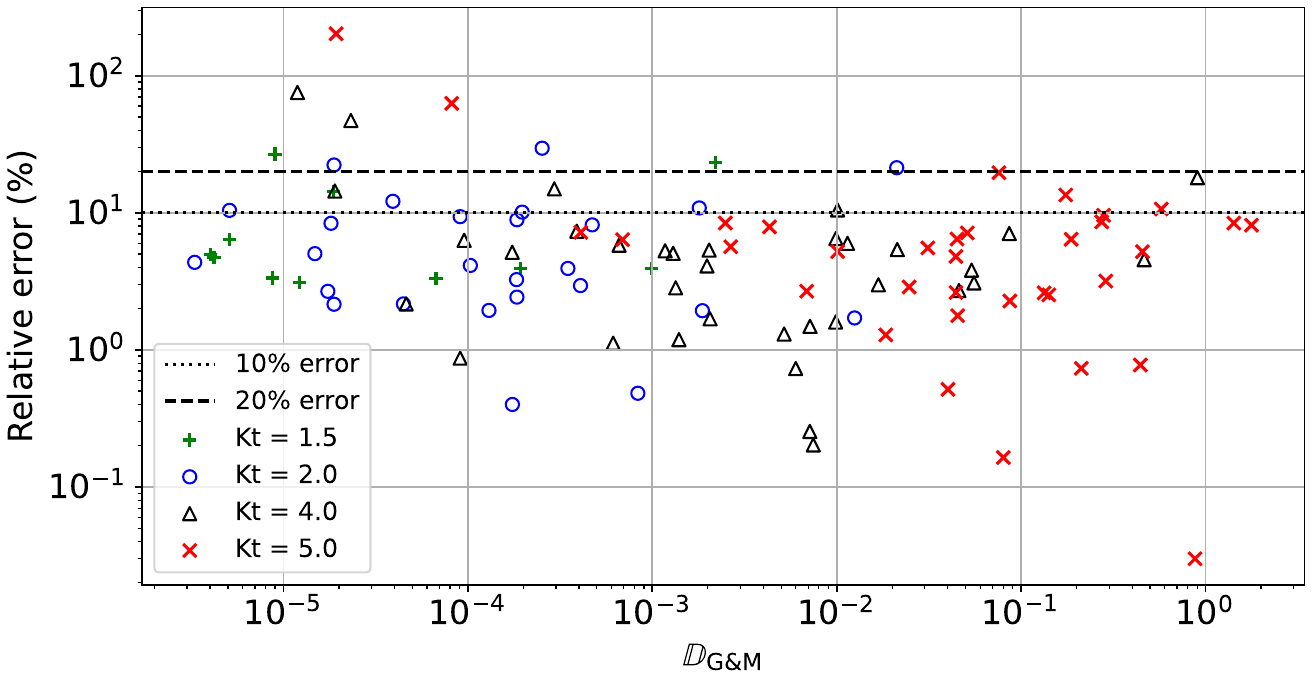}
    \caption{Scatter plot of the relative error of $\mathbb{D}_{\text{G\&M}}$ as a function of its ground truth value, for every point in the test set.}
    \label{fig:mre_blocks}
\end{figure}

%The outputs of both MLP models ($\overline{\mathscr{D}_\text{GAG}}$ and $\overline{\mathscr{D}_\text{G\&M}}$) are fatigue damage, which can lead to instantaneous breakage.
 Now, recall that we are using the mean absolute error (MAE) as the loss function in both cases
\begin{equation}
    \mathscr{L}_{\text{MAE}} = \frac{1}{m} \sum_{i=1}^{m} \left| y_{i} - \hat{y}_i \right|,
    \label{eq:mae}
\end{equation}
 where $m$ are the number of samples on the mini batch and $y_{i}$, $\hat{y}_i$ are the true and predicted damages, respectively. By construction, this loss function already penalizes error for higher damages than for lower damages. To confirm this
 % As mentioned before, these models predict the expected respective damage per flight, $\hat{\bar{D}}_{GAG}$ and $\hat{\bar{D}}_{G\&M}$. Thus, predictions are postprocessed to obtain accumulated damage, $\hat{D}_{GAG} = \hat{\bar{D}}_{GAG} \cdot n$ and $\hat{D}_{G\&M} = \hat{\bar{D}}_{G\&M} \cdot n$. 
 dependence of the prediction's error on the actual damage, in \autoref{fig:mre_GAG} and \autoref{fig:mre_blocks} we plot, in log-log scales, the relative error for the prediction of accumulated damages $\mathbb{D}_{\text{GAG}}$ and $\mathbb{D}_{\text{G\&M}}$ as a function of their ground truth values. Each point in these two scatter plots represents a $\{\text{Mission}, \text{PSE}, k_{t}\} \in \text{Test set}$. Clearly, model's performance is better for higher damages (more critical situations), and it is in the region of low damages that the error outliers tend to concentrate  (observe that the plots are logarithmic). This bias is even stronger for the GAG damage predictions --which is actually the main source of fatigue damage of the studied aircraft--, showing a statistically significant weak inverse correlation between the accumulated damage and the relative error of the prediction (Spearman's $\rho= -0.353$ with p-value 
 $=6.7 \times 10^{-5}$). In contrast, for the accumulated damage caused by gusts and maneuvers, the correlation is very weak and not statistically significant (Spearman's $\rho= -0.17$ with p-value $= 0.085$). However, the distribution outliers (e.g., following Tukey fences, those points with relative error larger than $Q3+1.5IQR=17.3 \%$) are mainly located in the low-damage range.
 
 % both in both cases a statistically significant inverse correlation between accumulated damage and relative error is observed. Spearman's $\rho= -0.44,-0.16$ for the GAG and G\&M cases, respectively, with p-values$=4.8 \times 10^{-7}, 0.11$).

%After modeling both sources of fatigue damage, $f_2$ in \autoref{fig:Pipeline} has been obtained. Since $f_3$ is simply \autoref{eq:Fatigue life}, fatigue life predictions, $N$, can finally be obtained.

\subsection{Phase III: Fatigue life results}
\label{sec:res_life}

\begin{figure}[h]
    \centering
    \includegraphics[width=0.8\linewidth]{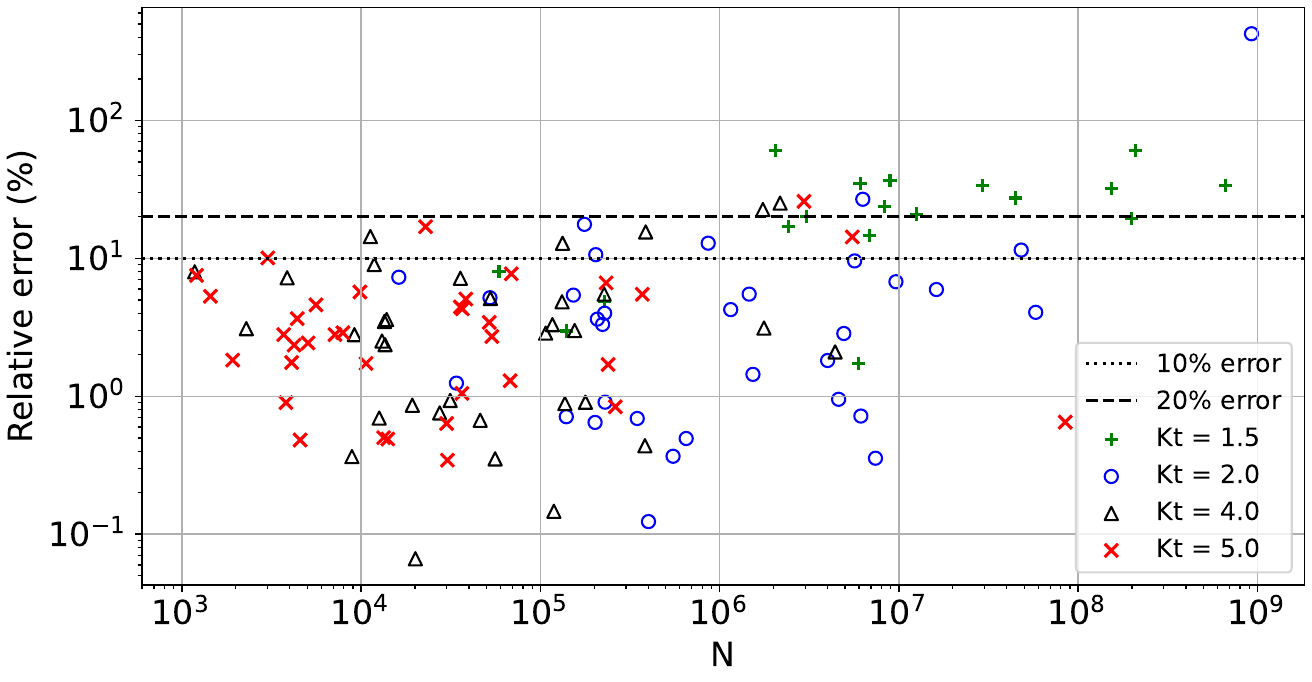}
    \caption{
    Scatter plot of the relative error of fatigue life prediction as a function of its ground truth value, for every point in the test set.}
    \label{fig:mre_N}
\end{figure}

Miner's Rule is finally applied to obtain total fatigue damage after the $n$ flights, and with Eq.\ref{eq:Fatigue life} fatigue life $N$ can easily be calculated. 
Since the MLP models for $f_2^{\text{GAG}}$ and $f_2^{\text{G\&M}}$ showed poorer performance for lower fatigue-related damages, errors propagate and thus we expect to get less accurate predictions of fatigue life for higher fatigue lives, since $N \propto \mathbb{D}^{-1}$.\\
In \autoref{fig:mre_N} we scatter plot the relative error of each fatigue life prediction $N$ as a function of its ground truth values. 
As observed, errors of the predictions are not uniformly distributed along fatigue life. For lives over $10^{6}$ the model shows  poorer performance.  This behavior is similar to that of the inverse of \autoref{fig:mre_GAG}. This is expected as the main source of damage for the studied aircraft lies in the GAG cycle.  This poorer performance for lives over $10^{6}$ flights does not represent a significant limitation for our model, since these very long lives do not represent critical situations and big errors do not pose a safety hazard. In contrast, shorter lives are more critical, as failure will be reached earlier in the aircraft operation.

\begin{figure}[h]
    \centering
    % Primera subfigura
    \begin{subfigure}{0.49\textwidth}
        \centering
        \includegraphics[width=1\textwidth]{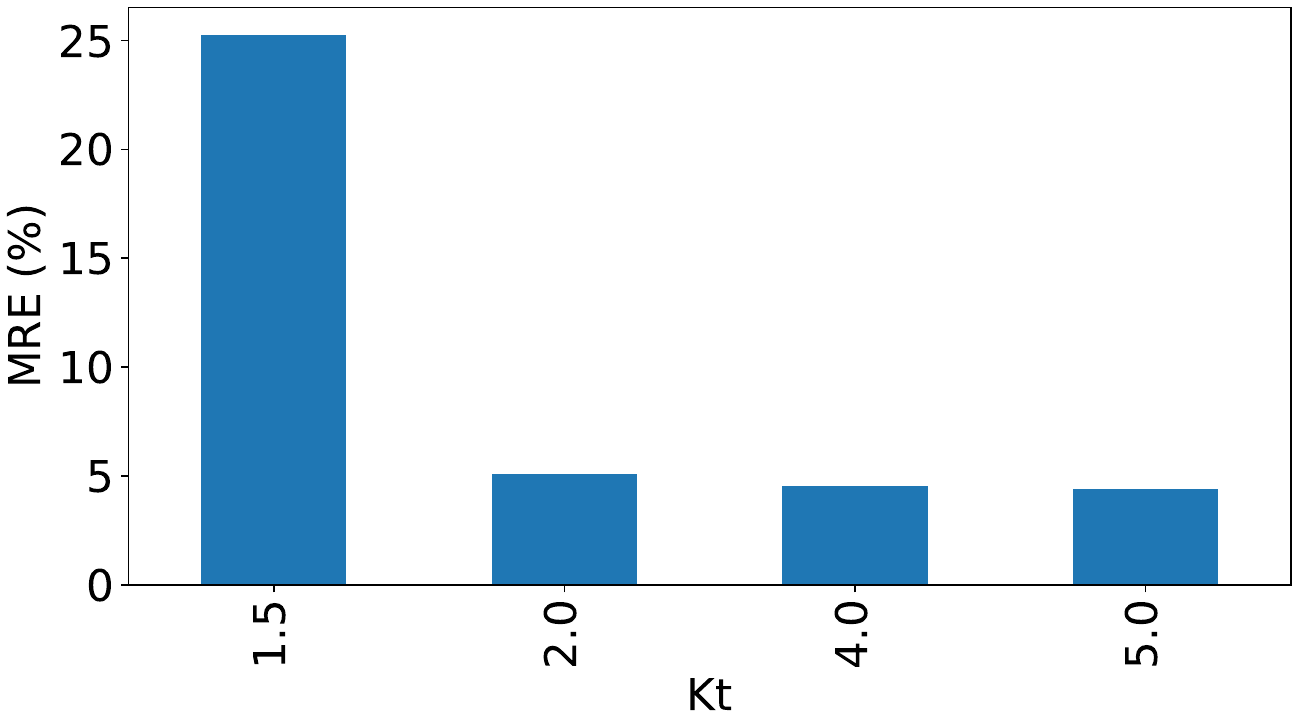}
        \caption{MRE of fatigue life prediction $N$, for all test set data grouped by their $k_{t}$ values.}
        \label{subfig: life_Kt}
    \end{subfigure}
    % Segunda subfigura
    \begin{subfigure}{0.49\textwidth}
        \centering
        \includegraphics[width=1\textwidth]{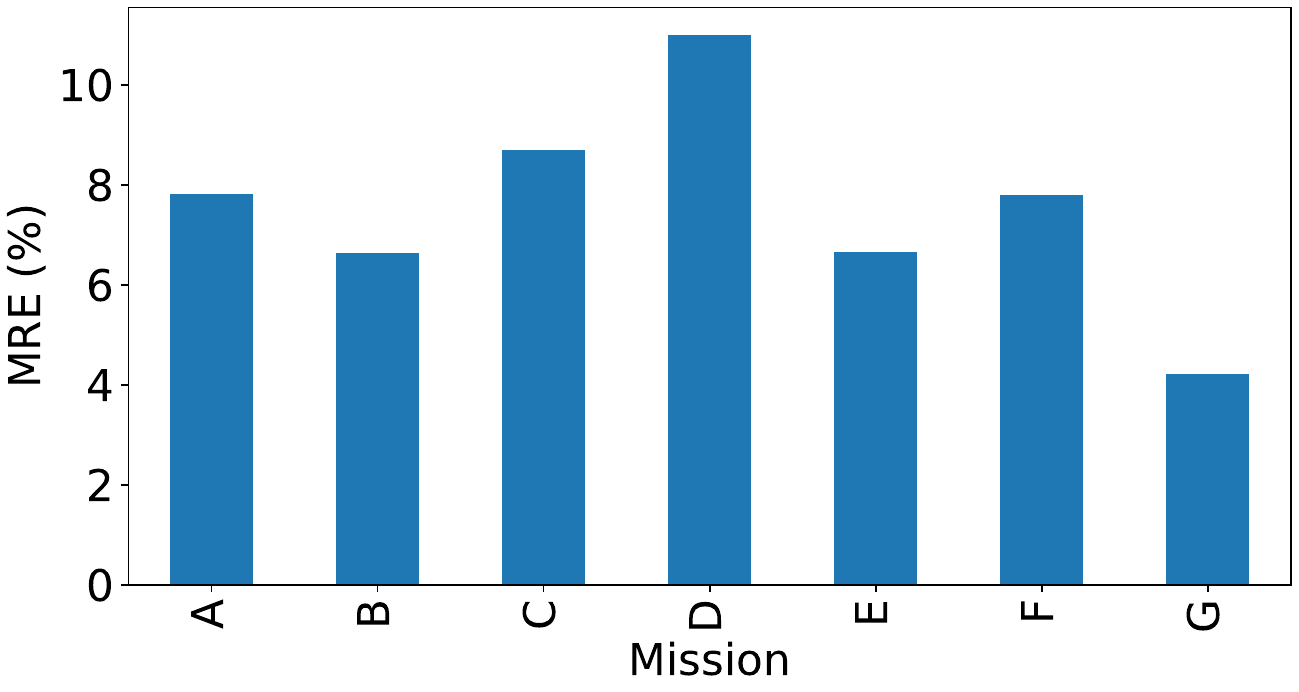}
        \caption{MRE of fatigue life prediction $N$, for all test set data grouped by their mission class.}
        \label{subfig: life_mission}
    \end{subfigure}
    % Tercera subfigura
    \begin{subfigure}{0.8\textwidth}
        \centering
        \includegraphics[width=0.8\textwidth]{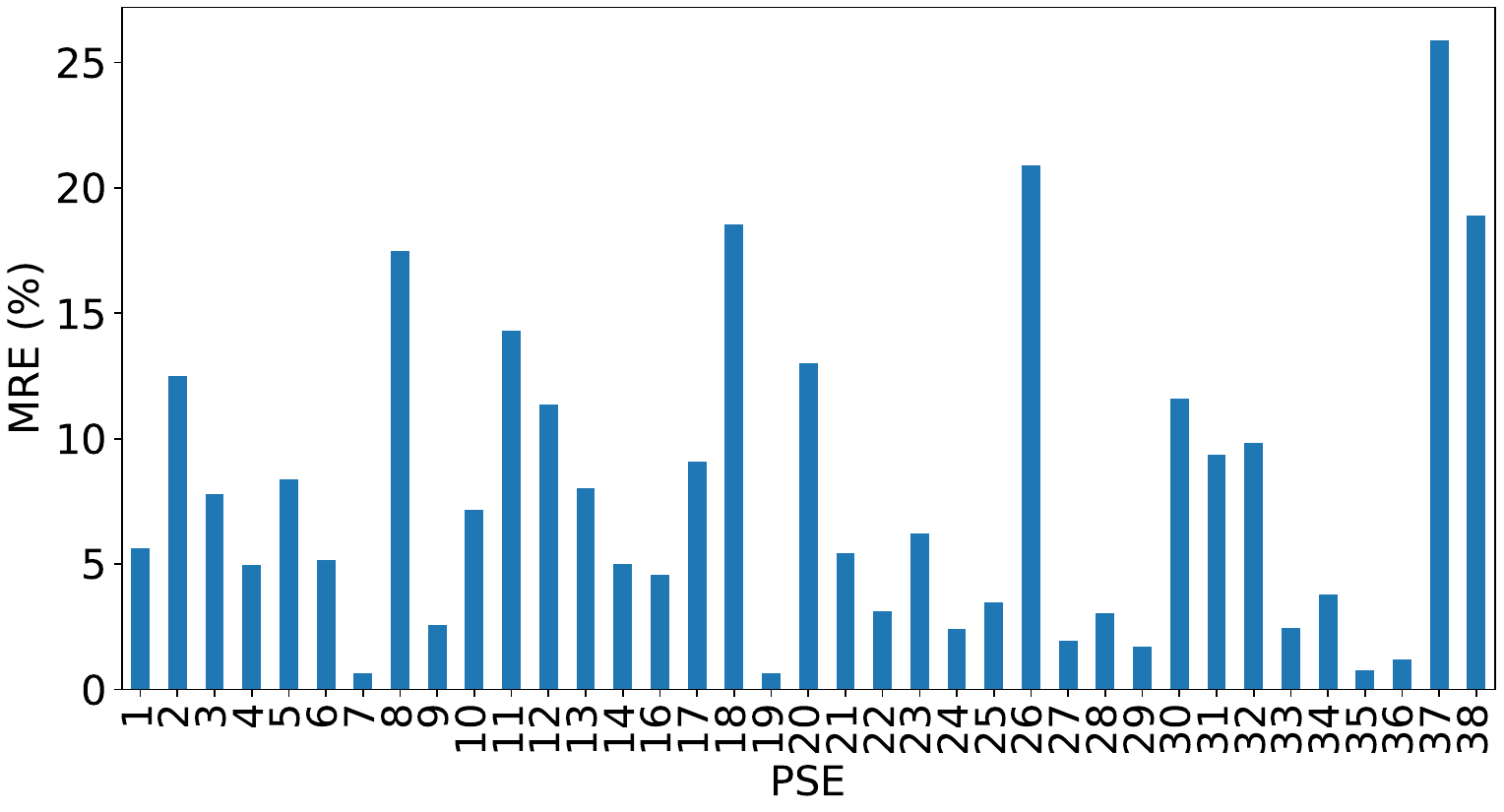}
        \caption{MRE of fatigue life prediction $N$, for all test set data grouped by PSEs.}
        \label{subfig:life_PSEs}
    \end{subfigure}

    % Caption general
    \caption{Mean relative error for the fatigue life prediction $N$, grouped by (a) $k_{t}$, (b) mission class, and (c) PSEs.}
    \label{fig: life_results}
\end{figure}

\medskip \noindent 
For completeness, in \autoref{fig: life_results} we also show the mean relative error of fatigue life prediction conditioned on $k_t$, mission and PSE, respectively. 
As observed in \autoref{subfig: life_Kt}, cases with smaller $k_t$ concentrate larger relative error. This is a sensible result: small $k_t$ values in general tend to have longer fatigue lives (i.e. the region where the ML model performs worse) because in these cases the local maximum stress is barely amplified by defects or geometry characteristics.

Likewise, in \autoref{subfig:life_PSEs} we observe that some PSEs show poorer fatigue life prediction, but this is simply because these are subjected to lower stresses, leading to higher fatigue lives and thus poorer model's performance.
%Hence, these SSEs are located in the region of poorer performance of the model. 
Incidentally, note that PSE 15 was not included in the plot as its high error was off-scale.  
%However, as explained earlier, this is not a big drawback for our model. 
Finally, \autoref{subfig: life_mission} shows only slight variability in performance across missions.

\medskip \noindent 
Finally, \autoref{tab:mre_N} depicts a summary of the main relative error statistics for fatigue life predictions and in \autoref{tab:mre_N_usage_region}, the same statistics are presented excluding samples with lives exceeding  $10^{6}$ flights, since these are non-critical and some of them can highly perturb the error statistics. In this latter region, performance is highly improved and predictions become more accurate. 

\begin{table}[h!]
\centering
\begin{minipage}{0.48\textwidth}
\centering
\caption{Full domain}
\label{tab:mre_N}
\begin{tabular}{cc}
\toprule
Relative error stats  & Values\\
\midrule
Mean                          & 11.37 \\
Standard deviation            & 39.48 \\
First quartile (25\%)         & 1.05 \\
Median (50\%)                 & 3.60 \\
Third quartile (75\%)         & 9.02 \\
Min error                     & 0.07 \\
Max error                     & 424.89 \\
\bottomrule
\end{tabular}
\end{minipage}
\hfill
\begin{minipage}{0.48\textwidth}
\centering
\caption{$N \in (10^3, 10^6)$}
\label{tab:mre_N_usage_region}
\begin{tabular}{cc}
\toprule
Relative error stats  & Values\\
\midrule
Mean                          & 3.99 \\
Standard deviation            & 4.03 \\
First quartile (25\%)         & 0.88 \\
Median (50\%)                 & 2.88 \\
Third quartile (75\%)         & 5.34 \\
Min error                     & 0.067 \\
Max error                     & 17.65 \\
\bottomrule
\end{tabular}
\end{minipage}
\end{table}

\begin{figure}[h!]
    \centering
    \includegraphics[width=1\linewidth]{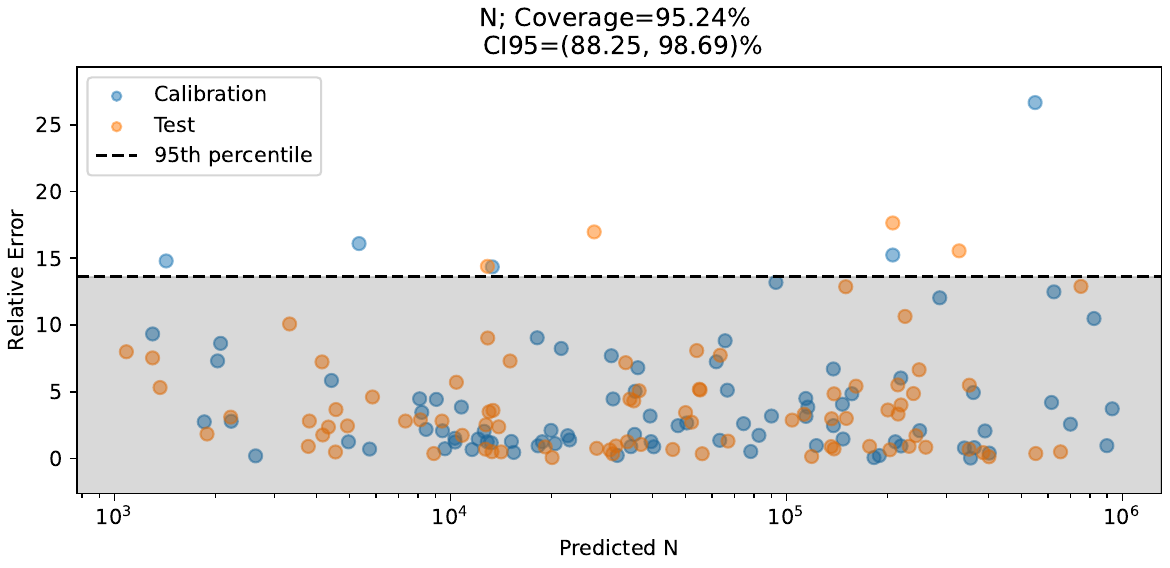}
    \caption{Prediction interval for the relative error}
    \label{fig:uncertainty_interval}
\end{figure}

\subsection{Uncertainty quantification: prediction intervals}
Since our ML pipeline aims to provide a solution for industrial and safety-critical applications, authorities require not only accuracy, but confidence in its predictions. In fatigue analysis, it us customary to use scatter factors to account for the variability in material fatigue properties and the uncertainties inherent in predicting fatigue life. 
To that end, following the principles of \cite{lacasa2025certification} we now build prediction intervals for fatigue life predictions.\\
We start by considering the relative error of the predictions of the calibration set in the region  $ N \in (10^3-10^6)$ to avoid impractical situations that can perturb the error statistics. Since the relative error is larger or equal to zero, we construct bootstrapped confidence interval of 95th percentile (CI95) of this metric  \cite{lacasa2025certification}.
In \autoref{fig:uncertainty_interval} we plot the bootstrap average of this percentile, such that the range $[0,\overline{P95}]$ in principle covers roughly the relative error of about $95\%$ of the points in the calibration and test set. The analysis showed that, using the validation set described in \autoref{sec:met_data_split} as the calibration set to build a prediction interval for the 95\% coverage, we achieve a coverage in the test set of 95.24\%, CI95 [88.25, 98.69]. This corresponds to an upper bound of the relative error of 13.50\%.\\
Consequently, when predicting the life of a new, previously unseen sample, the true fatigue life can be assured to be $N \in [\hat{N}(1-\varepsilon), \hat{N}(1+\varepsilon)]$ --- where $\varepsilon = 0.135$, and $\hat{N}$ is the predicted fatigue life --- with approximately 95\% confidence.

\subsection{A posteriori training/test split analysis}
Before assessing the performance of the different ML models, it is a good practice to check whether the training/test data split is adequate, as incorrect splits can be a source of p-hacking --which misleadingly amplifies the performance on the test set-- or poor generalization \cite{lacasa2025certification}. Now, since in this work the output data of model $f_1$ is the input data of model $f_2$, we now perform such data split analysis a posteriori, and focus in the input data of MLP models $f_2$ and $f_3$, as the configuration space is smaller than the one for $f_1$ (note that this analysis was performed for all steps):
\[ N = f_3(n,\overline{\mathscr{D}_{\text{GAG}}},\overline{\mathscr{D}_{\text{G\&M}}})  = f_3(n, f_2(\bar{\textbf{S}}_{flight}, \bar{\textbf{S}}_{ground}, t_{flight}, t_{ground}, k_t, n)), \]
where $f_2=\{f_2^{\text{GAG}},f_2^{\text{G\&M}}\}$.
Following \cite{lacasa2025certification}, we consider a split to be adequate if (i) finding the same data samples appearing in both the train and test sets has vanishingly small probability, (ii) the feature vector distributions of both sets are similar, and (iii) the test set data is adequately close from the training set.\\
Criterion (ii) is checked by performing 2-sample hypothesis tests (2-sample Kolmogorov-Smirnov (KS) \cite{scholz1987k}, 2-sample Anderson-Darling (AD) \cite{hodges1958significance} and 2-sample chi-2 tests) on marginal training and test distributions for each input variable. Criterion (iii) is implemented in the so-called tesselation and proximity method \cite{lacasa2025certification}, by detecting  test set points that are either excessively close (statistically speaking) to the training set (source of p-hacking), or statistically too far (isolated points with poor generalization).

%We will leverage the validation library introduced in \autoref{sec:met_ML} \cite{lacasa2025certification}, which includes algorithms and tests to answer these questions. This library was applied to detect p-hacking or isolated points in the test set- i.e., points that are either excessively close to or far from training points in the input space. In addition, two-sample versions of statistical tests such as Kolmogorov-Smirnov \cite{scholz1987k} or Anderson-Darling \cite{hodges1958significance} help to assess distribution similarity between two datasets. 

%\medskip \noindent 
%According to \autoref{eq:Fatigue life}, the input space for fatigue life predictions is: 
%\[ N = f_3(n,\textbf{D})  = f_3(n, f_2(\bar{\textbf{S}}_{flight}, \bar{\textbf{S}}_{ground}, t_{flight}, t_{ground}, k_t, n)). \]
%Based on these variables, the analysis is addressed. 

\begin{figure}[h!]
    \centering
    \includegraphics[width=1\linewidth]{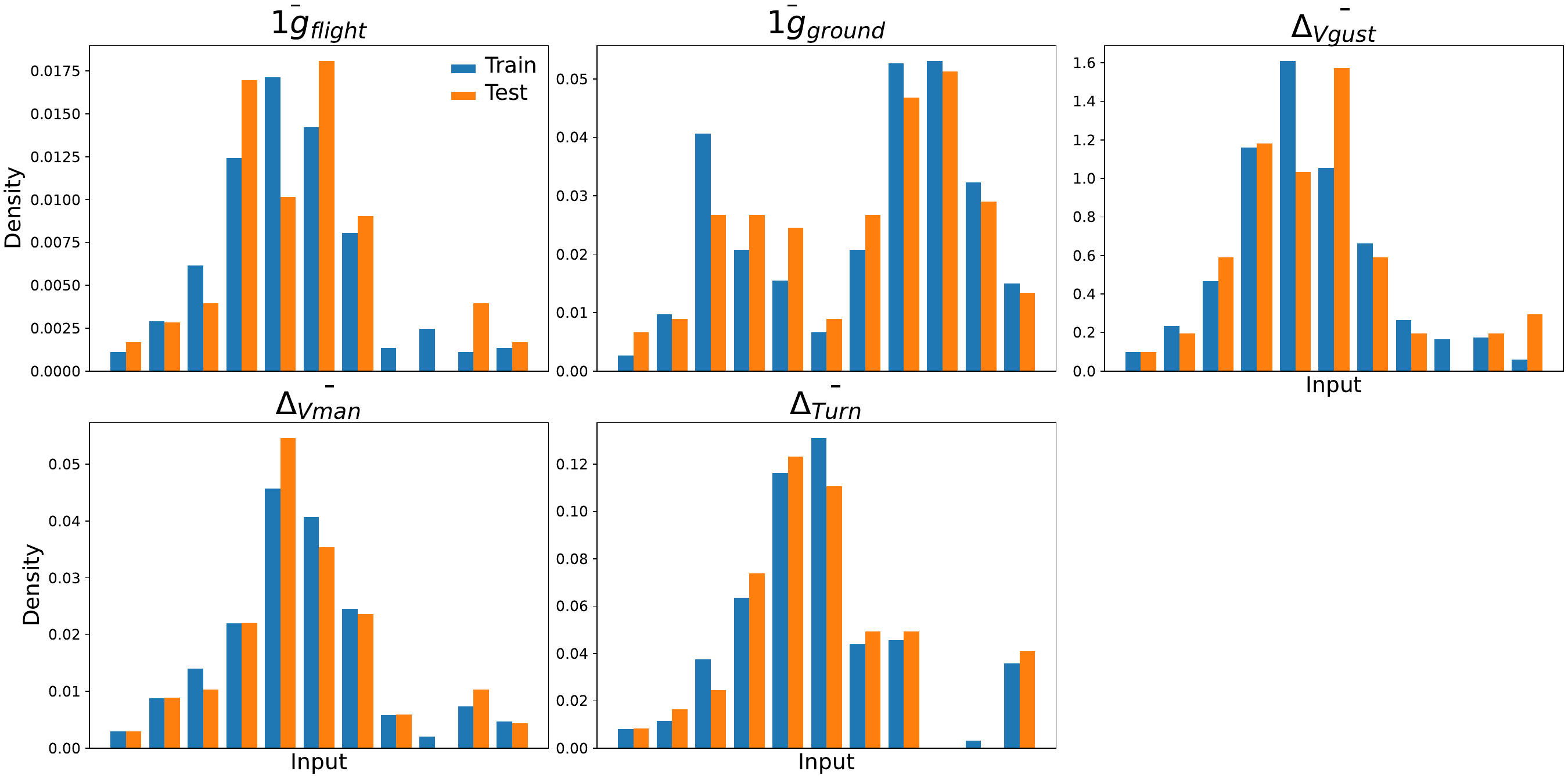}
    \caption{Training vs test marginal distributions for each numerical input variable.}
    \label{fig:input_num_histogram}
\end{figure}

\begin{figure}[h!]
    \centering
    \includegraphics[width=0.8\linewidth]{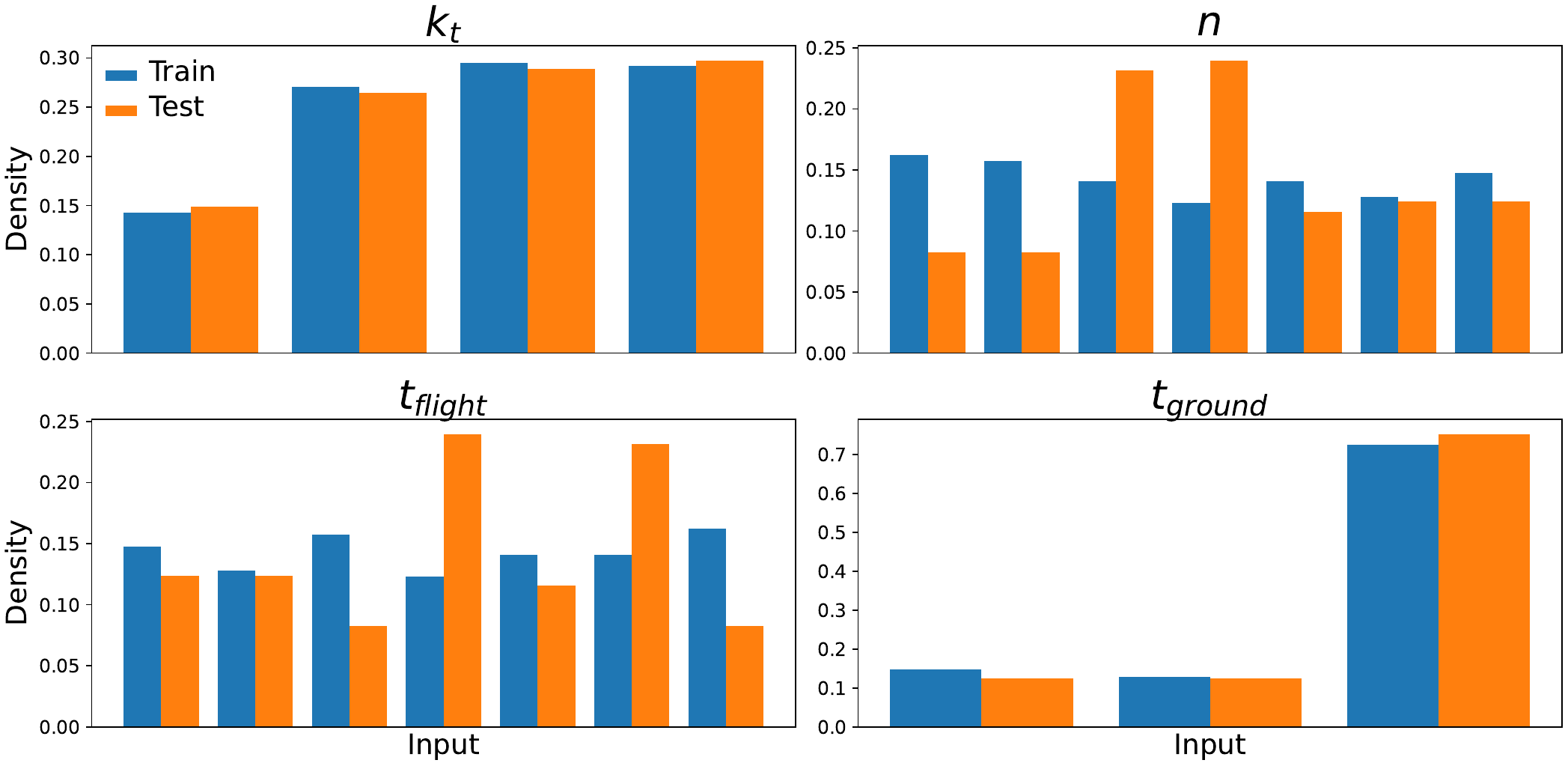}
    \caption{Training vs test marginal distributions for each categorical input variable.}
    \label{fig:input_cat_histogram}
\end{figure}

\begin{figure}
    \centering
    \includegraphics[width=0.8\linewidth]{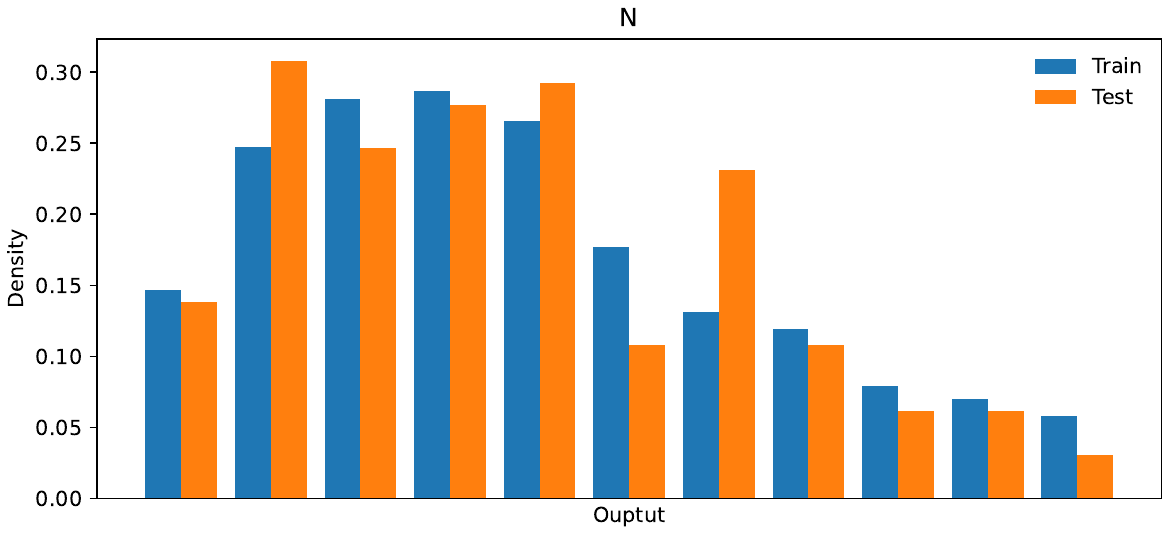}
    \caption{Training vs test marginal distributions for the output variable $N$.}
    \label{fig:output_dist}
\end{figure}

\medskip \noindent
In \autoref{fig:input_num_histogram} we plot the marginal training and test set distributions for each numerical input variable, whereas in \autoref{fig:input_cat_histogram} we plot the ones for the categorical variables. Although variables in \autoref{fig:input_cat_histogram} are actually numerical, for the sake of this analysis they are treated as categorical as they only take a small set of possible values. Therefore, they cannot be considered continuous and a different test is applied ($\chi^2$). Likewise, \autoref{fig:output_dist} presents the marginal training and test distributions for the output variable $N$. Visually, the training and test distributions are similar, although small differences exist. To quantify the significance of these, we have run several hypothesis tests, where the null hypothesis H0 is that, for each variable, each pair of populations are drawn from the same distribution. Since each hypothesis test has its own null hypothesis, no correction for multiple testing is required, and we adopt the standard choice for rejecting the null.
In \autoref{tab:dist_comparisons} we present the p-values of these tests, where the null hypothesis cannot be rejected for any variable except for $n$ and $t_{flight}$. Note nonetheless that, as observed in \autoref{fig:input_cat_histogram}, the range values for these two variables is similar in both sets, indicating that at least there are no isolated values in $n$ or $t_{flight}$ in the test set. 

Besides, from the 121 samples in the test set, the proximity method flagged three points (2.48\% of the test set) as p-hacking and eight as isolated samples (6.6\% of the test set). As expected, the p-hacking points showed overly optimistic predictions, with a mean relative error of 0.76\%, whereas this value for the isolated points is 10.21\%.

Finally, we can conclude that the test set is a good representation of the training set in both the input and output spaces and that the split is adequate, hence providing further assurance that the error and uncertainty quantification developed in previous sections represents well the performance of the model.

\begin{table}[h!]
\centering
\small
\begin{tabular}{@{}lcccccccccc@{}} % elimina espacio lateral extra
\toprule
\textbf{Test} & $k_t$ & $\bar{1g}_{flight}$ & $\bar{1g}_{ground}$ & $\bar{\Delta_{Vgust}}$ & $\bar{\Delta_{Vman}}$ & $\bar{\Delta_{Turn}}$ & $n$ & $t_{flight}$ & $t_{ground}$ & $N$ \\
\midrule
\text{KS}      & -      & 0.49   & 0.25 & 0.08 & 0.89 & 0.38 &         -       &  -   & -  & 0.95 \\
\text{AD}      & -      & 0.25   & 0.25 & 0.25 & 0.25 & 0.25 &         -       &   -  & -  & 0.25 \\
\textbf{$\chi$}² & 1.00   &  -     &   -  &   -  &  -   &   - &   \textcolor{red}{0.001} & \textcolor{red}{0.001} & 0.92 & - \\
\bottomrule
\end{tabular}
\caption{p-values for the two-sample Kolmogorov-Smirnov, Anderson-Darling and $\chi^2$ tests, comparing the empirical distribution of each input and output variables in the training vs test sets. Results indicate that, except for $n$ and $t_{\text{flght}}$, for the rest of input variables the null hypothesis of equal distribution cannot be rejected.}
\label{tab:dist_comparisons}
\end{table}

\newpage
\section{Conclusions}
\label{sec:Conclusions}
This study presents a modular machine learning-based pipeline for fatigue life prediction at different wing locations, across multiple stress concentration factors (\(k_t\)), using mission-level flight parameters as input. The framework offers a fast, scalable, and accurate complement to traditional simulation-based approaches, with direct applications in early-stage aircraft design, mission planning, and maintenance strategies.
The pipeline combines expert-domain feature engineering with deep learning models tailored to flight and ground segments. Once trained, it enables instant fatigue life and damage predictions which turn to be particularly accurate in the medium-to-low fatigue life regime, which is the critical regime for design decisions. The use of predicted stress components as intermediate variables supports both accuracy and physical interpretability. The pipeline has been statistically validated, making it consistent with ongoing initiatives to create certifiable ML tools in safety-critical aerospace systems.\\
Looking ahead, while the current study is limited to wing components due to data availability, the methodology is extensible to other materials and aircraft structures, such as the fuselage and stabilizers. Future work should also focus on broader structural coverage, improved stress modeling, modifications to the pipeline aimed at promoting conservativeness, (e.g., using modified loss functions to penalize non-conservative predictions), and further integration of uncertainty quantification conditioned on input space \cite{lacasa2025certification}.\\
In summary, the proposed approach represents a novel and effective contribution to fatigue life modeling, demonstrating the feasibility of using machine learning to complement and accelerate conventional analysis pipelines in aerospace structural integrity.\\

\noindent {\bf Acknowledgments --}
The authors acknowledge funding from project TIFON (PLEC2023-010251) funded by MCIN/AEI/10.13039/501100011033, Spain. LL acknowledges partial support from projects MISLAND (PID2020-114324GB-C22), project CSxAI (PID2024-157526NB-I00) funded by MICIU/AEI/10.13039/501100011033/FEDER, UE, project MdM Seal of Excellence (CEX2021-001164-M) funded by  MICIU/AEI/10.13039/501100011033 and from the European Commission Chips Joint Undertaking project No. 101194363 (NEHIL).
%The authors acknowledge funding from the European Union (project HERFUSE) under GA No 101140567. Views and opinions expressed are however those of the author(s) only and do not necessarily reflect those of the European Union or Clean Aviation Joint Undertaking. Neither the European Union nor Clean Aviation JU can be held responsible for them.
%LL acknowledges partial financial support from via project MISLAND
%(PID2020-114324GB-C22)
%and the Maria de Maeztu Program for units of Excellence (grant CEX2021-001164-M), both funded by MICIU/AEI/10.13039/501100011033, and funding from the European Commission Chips Joint Undertaking project No. 101194363 (NEHIL).
%GR acknowledges partial financial support received by the Grant DeepCFD (Project No. PID2022-137899OB-I00) funded by MICIU/AEI/10.13039/501100011033 and by ERDF, EU. 
Finally, all authors gratefully acknowledge the Universidad Politécnica de Madrid (www.upm.es) for providing computing resources on Magerit Supercomputer.

%% The Appendices part is started with the command \appendix;
%% appendix sections are then done as normal sections

\appendix
\section{Additional details for Phase I}
\label{appendix:PhaseI}
%In this appendix, we present the database, the split, the input-output variables and models of phase I of the pipeline.
%Missions are divided into operational segments such as taxi, climb, cruise, descent, and approach. As mentioned in \autoref{sec:met_prob}, this first phase aims to predict the stress vector \(\textbf{S}\) for each SSE and mission segment. 

\subsection{Flight parameters and stress vectors}\label{sec:PhaseI_database}
\noindent Each segment is parametrized by 12 scalar flight parameters, collected in a vector \(\mathbf{P} \in \mathbb{R}^{12}\) whose element description is detailed in
%\[
%\mathbf{P} = 
%\left\{
%\begin{aligned}
%\text{Flaps},\ \text{TAS},\ \text{Altitude},\ \text{Time},\ \text{Distance},\ \text{Thrust},\\ 
%\text{Pressure},\ \text{Mass},\ \%\text{CMA},\ \text{ZFW},\ \text{PL},\ \text{FW}
%\end{aligned}
%\right\}.
%\]
%The physical meaning of each variable in $\mathbf{P}$ is detailed in 
\autoref{tab:segment_variables}. We will use these parameters to raise the input of this phase. On the other hand, for each of the 38 wing locations (PSEs), stress variables are collected in a four-dimensional stress vector $\textbf{S}=(1g,\ \Delta{\text{vman}},\ \Delta{\text{vgust}},\ \Delta{\text{turn}})$, which represents the local stress at the component for different loading conditions. The physical meaning of each variable in $\mathbf{S}$ is detailed in \autoref{tab:segment_stresses}. Note that for ground segments $\Delta{\text{vman}} = \ \Delta{\text{vgust}} =  \Delta{\text{turn}} = 0$. 
%\end{itemize}
% Besides, these flight parameters induce certain stresses at each SSE, so for every mission segment the stress vector \textbf{S} is available for the 38 SSEs:
%\begin{itemize}
%   \item Stresses: These segment parameters \textbf{P} induce certain stresses at each SSE. Consequently, for every mission segment the stress vector \textbf{S} is available for the 38 SSEs. This is a 4-dimensional stress vector \(\mathbf{S} \in \mathbb{R}^4\), which represents the local stress at the component for different loading conditions:
%\[
% \mathbf{S} = \{1g,\ \Delta_{\text{Vman}},\ \Delta_{\text{Vgust}},\ \Delta_{\text{Turn}}\}.
% \]
% The physical meaning of each variable in $\mathbf{S}$ is detailed in \autoref{tab:segment_stresses}. Note that for ground segments $\Delta_{\text{Vman}} = \ \Delta_{\text{Vgust}} =  \Delta_{\text{Turn}} = 0$ 
%\end{itemize}

\begin{table}[h!]
\centering
\begin{tabular}{p{3cm} p{10cm}}
\toprule
Parameter & \textbf{Description} \\
\midrule
Flaps & Degree of extension of the flaps in the segment. \\
TAS & True airspeed during the segment. \\
Altitude & Altitude at which the segment takes place. \\
Time & Time spent in the segment. \\
Distance & Distance covered in the segment. \\
Thrust & Engine thrust during the segment. \\
Pressure & Pressure difference at the segment altitude. \\
Mass & Aircraft mass during the segment. \\
\%CMA & Relative position of the aircraft center of gravity with respect to the mean aerodynamic chord. \\
ZFW & Zero fuel weight of the aircraft. \\
PL & Payload weight. \\
FW & Fuel weight. \\
\bottomrule
\end{tabular}
\caption{Description of the 12 scalar flight parameters \(\mathbf{P} \in \mathbb{R}^{12}\) for each segment}
\label{tab:segment_variables}
\end{table}

\begin{table}[h!]
\centering
\begin{tabular}{p{4cm} p{9cm}}
\toprule
\textbf{Stress Component} & \textbf{Description} \\
\midrule
\(1g\) & Equilibrium stress at the studied PSE corresponding to steady flight. \\
\(\Delta_{\text{Vman}}\) & Incremental stress caused by a 1.5g vertical maneuver at the studied PSE. \\
\(\Delta_{\text{Vgust}}\) & Incremental stress caused by a 1.5g vertical gust at the studied PSE. \\
\(\Delta_{\text{Turn}}\) & Incremental stress caused by a 1.5g turn at the studied PSE. \\
\bottomrule
\end{tabular}
\caption{Description of the four stress components \(\mathbf{S} \in \mathbb{R}^4\) for each segment}
\label{tab:segment_stresses}
\end{table}

\subsection{Split} \label{sec:PhaseI_split}
\noindent As stated in \autoref{sec:met_data_split}, for each PSE, five missions were selected for training, one for validation and one for testing. However, in this phase, each \{segment, PSE\} constitutes a sample. This implies that, for each PSE, all the segments belonging to the five training missions are samples in the training set, and all the segments belonging to the validation and test missions belong to the validation and test sets respectively. This yields the split presented in \autoref{tab:phaseI_sizes}
\begin{table}[h!]
    \centering
    \begin{tabular}{cccc}
    \toprule
                                   & Train size & Validation size & Test size \\
         \midrule
         Flight segments           & 1644       & 310             & 344      \\
         Ground segments           & 380        & 76              & 76      \\
         \bottomrule
    \end{tabular}
    \caption{Set sizes for Phase I.}
    \label{tab:phaseI_sizes}
\end{table}

\subsection{Input, output variables and model architecture}
\noindent We shall distinguish in a separate way the input variables of $f_1^{\text{ground|PSE}}$ and $f_1^{\text{flight}}$. 
%In this phase, ground segments (taxi) and flight segments (climb, cruise, descent, approach) are studied separately. 
During ground segments (i.e. \texttt{taxi}) the wing is mainly subjected to gravity loads  (the weight of the wing itself, the engine and the weight of the fuel stored in the wings). Since the wing and engine are the same across all the missions, for each PSE only one feature is the input variable to the model $f_1^{\text{ground|PSE}}$: FW.\\
For flight segments (i.e. \texttt{climb}, \texttt{cruise}, \texttt{descent}, \texttt{approach}), wings are also subjected to aerodynamic loads, so more variables will impact on stresses (for instance, lift is proportional to \text{TAS}$^{2}$). So the following set of features are used as input variables for $f_1^{\text{flight}}$: \{PSE (binary encoded), \text{Flaps},\ \text{Altitude},\ \text{TAS},\ \text{Mass},\ \text{FW},\ \text{Thrust}\}. All of them were normalized with a regular MinMax Scaler.

%\begin{align*}
%&X_{\text{flight}} = SSE + \{\text{Flaps},\ \text{Altitude},\ \text{TAS},\ \text{Mass},\ \text{FW},\ \text{Thrust} \}_6 \\
%&X_{\text{ground}} = \{\text{Mass},\ \text{FW}\}_2
%\end{align*}

 %For flight segments, the aircraft location (categorical) and those flight parameters (numerical) that strongly influence the loads at the wing during the segment. For ground segments the SSE was not included since a single model was developed for each one. 
 
\medskip \noindent Likewise, the sole output variable of each $f_1^{\text{ground|PSE}}$ is $1g$, whereas for flight segments, the model outputs the equilibrium stress and the incremental stresses ($1.5 g's$) caused by vertical gusts, vertical maneuvers and turns at the PSE, i.e. a four-dimensional output vector $(1g, \Delta\text{vman}, \Delta\text{vgust}, \Delta\text{turn})$.
%\begin{align*}
%  &Y_{flight}=  \{ 1g, \Delta_{Vman}, \Delta_{Vgust}, \Delta_{Turn}\}_{4}\\
%  &Y_{ground} = \{1g\}_1
%\end{align*}
%For flight segments, the equilibrium stress and the incremental stresses ($1.5 g's$) caused by vertical gusts, vertical maneuvers and turns at the SSE in that mission segment. 
%For ground segments the equilibrium stress only. 

\medskip \noindent
Finally, the hyperparameters of the deep neural network used to approximate $f_1^{\text{flight}}$ and other optimization parameters are depicted in \autoref{tab: unitarios_MLP}.
%Ground segments exhibited a very simple implicit relationship between $Y_{ground}$ and $X_{ground}$. For these segments, a simple polynomial regression (second order) for each SSE was sufficient to yield accurate predictions. In contrast, flight segments showed a more complex input-output relationship  so a multilayer perceptron, MLP was adopted (architecture specified in \autoref{tab: unitarios_MLP}). Categorical variables were encoded (one-hot encoding), and all of them were normalized with a regular MinMax Scaler. 
Hyperparameter tuning was performed manually on the validation set to ensure generalization.

\begin{table}[h!]
\centering
\begin{tabular}{cc}
\toprule
Train/Val/Test size                   & 1664/310/344 \\  
Hidden layer sizes                    & 50/50 \\
Activation Function                   & ReLU \\
Loss Function                         & MAE  \\
Optimizer                             & Adam  \\
Weights initialization                & Xavier \\
Learning rate $\alpha$                &  $8 \cdot 10^{-3}$ \\ 
Epochs                                & 1000 \\
Regularization                        & None\\ 
Mini batch Size                       & 256 \\ 
Scheduler: $\gamma$, Step             & 0.975, 30 \\
\bottomrule
\end{tabular}
\caption{Hyperparameters and optimization choices for the deep MLP used to approximate the stress properties for any given PSE for flight segments ($f_1^{\text{flight}}$).}
\label{tab: unitarios_MLP}
\end{table}

\newpage
\section{Phase II: Additional details for Phase II}\label{appendix:PhaseII}
\subsection{Data preprocessing}\label{sec:Data_preprocessing}
%As mentioned in \autoref{sec:met_prob}, stress predictions \textbf{S} obtained in phase I are required to form the input for predicting damages \textbf{D} in phase II. 
\noindent Predicted stresses $\bf$ from all the mission segments are averaged to form $$\bar{\mathbf{S}}_{\text{ground}} = \langle 1g \rangle \in \mathbb{R}$$ for ground segments and $$\bar{\mathbf{S}}_{\text{flight}} = (\langle 1g\rangle_\text{flight}, \langle \Delta\text{vman} \rangle_\text{flight}, \langle \Delta\text{vgust} \rangle_\text{flight}, \langle\Delta\text{turn}\rangle_\text{flight}  )\in \mathbb{R}^4$$ for flight segments. In addition, the durations associated with each mission segment are used to compute $t_{flight}$ and $t_{ground}$. 
%\[
%\mathbf{D} = f_2(\bar{\mathbf{S}}_{\text{ground}}, \bar{\mathbf{S}}_{\text{flight}}, 
%t_{\text{flight}}, t_{\text{ground}}, k_t, n),
%\]
%where: 
%\[
%\bar{\mathbf{S}}_{\text{flight}} = \{\bar{1g}_{flight}, \bar{\Delta_{Vgust}}, \bar{\Delta_{Vman}}, \bar{\Delta_{Turn}}\}_4
%\] 
%\[
%\bar{\mathbf{S}}_{\text{ground}} = \{\bar{1g}_{ground}\}_1
%\]
 %contains the averaged equilibrium stress during ground segments (taxi). 
All the averages $\langle \cdot \rangle$ above are computed in a similar way:
%The same procedure is applied to compute all the averaged values in $\bar{\mathbf{S}}_{\text{flight}}$ and $\bar{\mathbf{S}}_{\text{ground}}$. 
as an example, the averaged equilibrium stress during flight segments $$\langle 1g\rangle_{flight}=\frac{1}{t_{flight}} \sum 1g_{i} \cdot t_{i},$$
%\begin{equation}
%        \bar{1g}_{flight} = \frac{1}{t_{flight}} \sum1g_{i} \cdot t_{i}, 
%    \end{equation}
where $1g_{i}$ and $t_{i}$ are the predicted equilibrium stress of flight segment $i$ and its duration, respectively.

\subsection{Database}
\noindent  Load cycles: For each segment, load cycles caused by vertical maneuvers, gusts, and turns are included. In addition, the $n $ GAG cycles corresponding to the $n$ flights are also included. As detailed in \autoref{sec:met_prob}, these are stored as a vector with three components: Maximum stress, Minimum stress and Number of occurrences (after \(n\) flights).  The associated damage for each of these cycles is also available (for the four $k_t$ values).
%\item Fatigue Damage and Life:
%\begin{itemize}
%    \item Fatigue damage values \(\mathbf{D}\), induced by the load cycles (gusts and maneuvers) and by the GAG cycle (for the four stress concentration factor $k_t$ values). 
%    \item Fatigue life \(N\), computed for four values of the stress concentration factor \(k_t\).
%\end{itemize}
%These are the accumulated quantities over the \(n\) flights recorded in the mission.

%\item Preprocessed data: $\bar{\mathbf{S}}_{\text{flight}}$, $\bar{\mathbf{S}}_{\text{ground}}$, $t_{flight}$ and $t_{ground}$. 
%\end{itemize}

\subsection{Split} \label{sec:PhaseII_split}
\noindent As stated in \autoref{sec:met_data_split}, for each PSE, five missions were selected for training, one for validation and one for testing (with their respective four $k_t$ values). In this phase, every \{mission, PSE, $k_t$\} represents a sample. After filtering extreme values and unrealistic data, the set sizes for the GAG model and the gusts and maneuvers model are presented in  \autoref{tab:phaseII_sizes}.
\begin{table}[h!]
    \centering
    \caption{Set sizes}
    \label{tab:phaseII_sizes}
    \begin{tabular}{cccc}
    \toprule
                                              & Train size & Validation size & Test size \\
         \midrule
         GAG                           & 621       & 125              & 123      \\
         Gusts and maneuvers           & 533       & 104              & 107      \\
         \bottomrule
    \end{tabular}
\end{table}

\subsection{$f_2^{\text{GAG}}$: input, output variables and model specs} \label{sec:GAG_inputs}

%}$\mathscr{D}_{\text{GAG}}$
\noindent $f_2^{\text{GAG}}$ takes 9 numerical input variables: $k_{t}, \bar{\mathbf{S}}_{\text{flight}} \in \mathbb{R}^4 , \bar{\mathbf{S}}_{\text{ground}}, 
n, t_{\text{flight}}, t_{\text{ground}}$. 
%In order to define the problem correctly and define the GAG cycle as accurately as possible with the data available, thereby facilitating the learning process of the model, the following input variables were raised. 
%\subsubsection*{Inputs}
%\[ \textbf{X} = 
%\left\{
%\begin{aligned}
%k_{t}, \bar{\mathbf{S}}_{\text{flight}} , \bar{\mathbf{S}}_{\text{ground}}, 
%n, t_{flight}, t_{ground}
%\end{aligned}
%\right\}_{9},
%\] where 
%\[  \bar{\mathbf{S}}_{\text{flight}} = \{\bar{1g}_{flight}, \bar{\Delta_{Vgust}}, \bar{\Delta_{Vman}}, \bar{\Delta_{Turn}}\}_4 \] and
% \[ \bar{\mathbf{S}}_{\text{ground}}= \{ \bar{1g}_{ground}\}_1 \] are the preprocessed averaged stresses from \autoref{sec:Data_preprocessing}. 
Note that as only wing PSEs are studied and all the wing PSEs have the same material, the material is not an input variable. The physical rationale behind each one of these input variables is as follows:
\begin{itemize}
    \item $k_{t}$: The stress concentration factor, \( k_t \), has a strong detrimental effect on fatigue life. This variable quantifies the amplification of local stress caused by defects or geometric characteristics, such as holes or joints. High $k_t$ values promote crack nucleation. It is defined as:
\begin{equation}\label{eq:Stress_concentration_factor}
    k_t = \frac{\text{Local maximum stress}}{\text{Nominal stress}}
\end{equation} 
and it is a key variable in this problem.
    \item $\langle 1g\rangle_{\text{ground}}$ and $\langle 1g\rangle_{\text{flight}}$: about the form and amplitude of the GAG cycle. 
    \item  $\langle {\Delta\text{vgust}}\rangle, \langle {\Delta\text{vman}}\rangle, \langle {\Delta\text{turn}}\rangle$: these averaged incremental stresses provide information about the magnitude of the stress oscillations caused by gusts and maneuvers. In the GAG cycle, the maximum stress will always be greater than $\langle 1g\rangle_{\text{flight}}$. The same happens with the minimum stress.
    \item $n$, $t_{flight}$, $t_{ground}$: These variables have a statistical impact. Longer flights increase the likelihood of encountering greater gusts or executing more demanding maneuvers, which augments the amplitude of the GAG cycle and, consequently the average GAG damage,  $\overline{\mathscr{D}_{\text{GAG}}}$, which is the output of the model. Besides, since the number of samples is finite, $n$ is included to account for potential deviations from the true mean damage. 
\end{itemize}
Input and output variables were normalized with MinMax Scaler and a logarithm was applied to the damage due to the scale difference between samples.\\
%\subsubsection*{Output}
%\[ Y= \{ \bar{D_{GAG}}\}_1 \]
%As detailed in \autoref{sec:met_prob}, the model will predict the average GAG damage, $\bar{D_{GAG}} $, which is easily postprocessed to obtain the accumulated GAG damage after $n$ flights, $D_{GAG} = \bar{D_{GAG}} \cdot n$
%\subsubsection*{Model}
Finally, the specs of the MLP model and its hyperparameters are shown in \autoref{tab: GAG_MLP}. Hyperparameter tuning was performed manually with the validation set to ensure generalization. 
% probe un par de veces optuna, pero me proponia redes más complejas y que no mejoraban los resultados

\begin{table}[h!]
\centering
\begin{tabular}{cc}
\toprule
Train/Validation/Test sizes           & 621/125/123 \\
Hidden layer sizes                    & 64/64 \\
Activation Function                   & tanh \\
Loss Function                         & MAE  \\
Optimizer                             & Adam  \\
Weights initialization                & Xavier \\
Learning rate $\alpha$                &  $8 \cdot 10^{-3}$ \\ 
Epochs                                & 5000 \\
Regularization                        & Dropout, 0\\ 
Mini batch Size                       &    128 \\ 
Scheduler: $\gamma$, Step             & 0.975, 30 \\
\bottomrule
\end{tabular}
\caption{Specifications and hyperparameters of the MLP of $f_2^{\text{GAG}}$.}
\label{tab: GAG_MLP}
\end{table}

\subsection{$f_2^{\text{G\&M}}$: input, output variables and model specs}
\noindent In this case we only have 7 input variables:
$k_{t},\bar{\mathbf{S}}_{\text{flight}}\in \mathbb{R}^4,
n, t_{flight}$. The rationale of each variable is similar to that for $f_2^{\text{GAG}}$; $\langle 1g\rangle_{\text{flight}}$ shows the stress level around which the oscillations that cause this damage take place. The magnitude of these oscillations is captured by  $\langle {\Delta\text{vgust}}\rangle, \langle {\Delta\text{vman}}\rangle, \langle {\Delta\text{turn}}\rangle$ and $n$, $t_{flight}$ have the same statistical impact. Besides, longer flights typically imply more gusts encountered and maneuvers executed (not necessarily larger) and consequently greater damage.  The output variable is the per-flight damage $\overline{\mathscr{D}_{\text{G\&M}}}$, such that the cumulated damage is simply $\mathbb{D}_{\text{G\&M}}=n\cdot \overline{\mathscr{D}_{\text{G\&M}}}$. 
%\[  \bar{\mathbf{S}}_{\text{flight}} = \{\bar{1g}_{flight}, \bar{\Delta_{Vgust}}, \bar{\Delta_{Vman}}, \bar{\Delta_{Turn}}\}_4 \]
%Seven numerical (scalars) variables. The physical rationale behind each one of them is now explained: 
%\begin{itemize}
%    \item $k_{t}$: Great impact on fatigue damage and life.
%    \item $\bar{1g}_{flight}$: Represents the baseline stress around which the stress fluctuations due to gusts and maneuvers occur. Higher values of this variable lead to higher damages. 
%    \item  $\bar{\Delta_{Vgust}}, \bar{\Delta_{Vman}}, \bar{\Delta_{Turn}}$: Provide information about the magnitude of the stress fluctuations during the mission, which are what cause this damage. 
%    \item $n$: Same rationale as for the GAG damage model. 
%    \item $t_{flight}$: A longer flight will necessarily lead to more gusts and maneuvers and consequently more fatigue damage.
%\end{itemize}
%\subsection*{Output}
%\[ Y= \{ \bar{D_{G\&M}}\}_1 \]
%As detailed in \autoref{sec:met_prob}, the model will predict the average damage, $\bar{D_{G\&M}} $, which is easily postprocessed to obtain the accumulated damage after $n$ flights, $D_{G\&M} = \bar{D_{G\&M}} \cdot n$
%\subsubsection*{Model}
Once again, variables were normalized with MinMax Scaler and logarithm was applied to the damage. The MLP specs and hyperparameters are specified in \autoref{tab: Blocks_MLP}. Hyperparameter tuning was performed manually with the validation set to ensure generalization. 

\begin{table}[h!]
\centering
\caption{MLP architecture}
\label{tab: Blocks_MLP}
\begin{tabular}{cc}
\toprule
Train/Validation/Test size            & 533/104/107 \\
Hidden layer sizes                    & 64/64 \\
Activation Function                   & ReLU \\
Loss Function                         & MAE  \\
Optimizer                             & Adam  \\
Weights initialization                & Xavier \\
Learning rate $\alpha$                &  $8 \cdot 10^{3}$ \\ 
Epochs                                & 5000 \\
Regularization                        & Dropout, 0.001\\ 
Mini batch Size                       &    128 \\ 
Scheduler: $\gamma$, Step             & 0.975, 30 \\

\bottomrule
\end{tabular}
\end{table}

\newpage
\section{Statistical validation of moment closure and finite sample deviations}
\label{appendix:Statistical_analysis}

As stated in Section \ref{sec:Methodology}, our method predicates on approximating the individual damages in each flight --the latent random variables $\mathscr{D}_{\text{GAG}}$ and $\mathscr{D}_{\text{G\&M}}$ -- by their sample mean. In this appendix, we give insight into the validity of the implicit assumptions behind this procedure with an analysis based on the GAG damage.\\
There are mainly two approximations: 

\begin{itemize}
\item First, we identify each realization of the GAG damage random variable $\mathscr{D}_{\text{GAG}}$ with the observed per-flight average $\overline{\mathscr{D}_{\text{GAG}}}$, i.e, we approximate a random variable with its (estimated) first moment. This is sometimes called a mean-field approximation or a first-moment closure, and it is a valid approximation when higher moments of the distribution (e.g. fluctuations of the random variable around its mean value) are relatively small (in particular, much smaller than the mean). Such comparison can be made by computing the so-called squared coefficient of variation, $\text{CV}^{2} = \frac{\sigma^{2}}{\overline{\mathscr{D}_\text{GAG}} ^{2}}$, where $\sigma^2$ and $\overline{\mathscr{D}_\text{GAG}}$ are the sample variance and sample mean of the GAG damage per-flight, respectively. First moment closures usually require $\text{CV}^2 \ll 1$. In \autoref{tab:moments} we provide an illustrative table with results for three missions for a given PSE and $k_{t}$. As observed, identifying each realization of the per-flight random variable $\mathscr{D}_\text{GAG}$ with its sample average is not fully justified yet, $\text{CV}^2 \le1$, not $\text{CV}^2 \ll 1$, suggesting non-negligible fluctuations. However, as specified in \autoref{sec:Methodology} the damage of one particular flight is not the variable of interest. On the contrary, the important variable is the cumulated damage after $n$ flights, $\mathbb{D}_\text{GAG} = \sum_{i=1}^n \mathscr{D}_{\text{GAG}}$. Accordingly, the first moment closure needs to hold with respect to this cumulated random variable $\mathbb{D}_\text{GAG}$. 
Now, for large $n$ (when expected values coincide with sample averages), the expected value of the cumulated GAG damage $\mathbb{E}(\mathbb{D}_\text{GAG})\equiv\overline{\mathbb{D}_\text{GAG}}$, which for ease of notation we label it  $\mathbb{D}_\text{GAG}$, reads
$\mathbb{D}_\text{GAG} = \overline{\mathscr{D}_\text{GAG}} \cdot n$, since 
\begin{equation}
    \mathbb{E}(\mathbb{D}_\text{GAG}) = \mathbb{E}(\sum_{i=1}^{n} \mathscr{D}_{\text{GAG}}) = \sum_{i=1}^{n} \mathbb{E}(\mathscr{D}_{\text{GAG}}) = n\overline{\mathscr{D}_{\text{GAG}}}.
\end{equation}
At the same time, as we are assuming flight independence we have
\begin{equation}
\text{Var}(\mathbb{D}_\text{GAG}) = \text{Var}(\sum_{i=1}^{n}\mathscr{D}_\text{GAG}) = \sum_{i=1}^{n}\text{Var}(\mathscr{D}_\text{GAG}) = n\cdot \sigma^{2}. 
\end{equation}
Accordingly, assessment of the first moment closure approximation via the squared coefficient of variation reads now
\begin{equation}
    \text{CV}^{2}_{\mathbb{D}_\text{GAG}} = \frac{\text{Var}(\mathbb{D}_\text{GAG})}{\mathbb{E}(\mathbb{D}_\text{GAG})^{2}} = \frac{n\cdot\sigma^{2}}{n^{2}\cdot \overline{\mathscr{D}_\text{GAG}}^{2}} = \frac{\text{CV}^{2}}{n}
\end{equation}
In short, the first moment closure is a substantially improved approximation for large $n$ (see \autoref{tab:moments} for three different missions), and overall the procedure is thus sound and justified.
%The deviations of the accumulated damage decrease with respect its expected value as $\frac{1}{\sqrt{n}}$. In \autoref{tab:moments} we present the coefficient of variation of the accumulated damage,  showing that our approximation is accurate for all the ranges of $n$. 
%\begin{table}[h!]
%\centering
%\caption{Sample moments for different $n$}
%\label{tab:moments}
%\begin{tabular}{c c c}
%\hline
%\textbf{$n$} & $CV^{2}$ & $CV_{D_{GAG, i}}$  & $CV^{2}_{\mathbb{D}_\text{GAG}}$ & $CV_{D_{GAG}}$\\
%\hline
%$\approx 800$  & 1.06 & 1.03   & 0.0013       & 0.036\\
% $\approx 800$& $5.9\times10^{-5}$ & $3.7\times10^{-9}$ & $2.2\times10^{-10}$ \\
%$\approx 1300$ &  0.27  &   0.52  & 0.00021  & 0.014 \\
% $\approx 1300$ & $6.19\times10^{-5}$ & $1.05\times10^{-9}$ & $3.53\times10^{-10}$ \\
%$ \approx 13000 $ & 0.091  & 0.30  &  0.0000069  & 0.0026 \\
% $\approx 13000$ & $4.26\times10^{-5}$ & $1.66\times10^{-10}$ & $1.98\times10^{-9}$ \\
%\hline
%\end{tabular}
%\end{table}
%\end{itemize} 
\begin{table}[h!]
\centering
\begin{tabular}{c c c}
\hline
\textbf{$n$} & $\text{CV}^{2}$  & $\text{CV}^{2}_{\mathbb{D}_\text{GAG}}$ \\
\hline
% $\approx 800$  & 1.06 & 0.0013 \\
$\approx 800$  & 1 & 0.001 \\
% $\approx 1300$ &  0.27  & 0.00021   \\
$\approx 1300$ &  0.3 & 0.0002  \\
% $ \approx 13000 $ & 0.091 & 0.0000069  \\
$ \approx 13000 $ & 0.09 & 0.000007 \\

\hline
\end{tabular}
\caption{Squared coefficients of variation for $\mathscr{D}_\text{GAG}$ and $\mathbb{D}_\text{GAG}$, for three different PSEs illustrating very different number of flights $n$.}
\label{tab:moments}
\end{table}
%\textbf{NOTA TEMPORAL: Para nosotros, esta desviación no existe. Me explico: 
%Nuestro ground truth es \textit{exactamente} $D_{GAG} =  \overline{\mathscr{D}_\text{GAG}}\cdot n$. Es decir un error de 0\% del MLP al predecir $\overline{\mathscr{D}_\text{GAG}}$ implica un error también del 0\% en el acumulado, $D_{GAG}$. Esto es por que el ground truth $\overline{\mathscr{D}_\text{GAG}}$ se sacó con un preproceso como $\frac{D_{GAG}}{n}$. Por lo tanto, mi conclusión es la siguiente: 
%Esta aproximación no es nuestra, si no de la metodología tradicional. La met.tradicional está diciendo que para n vuelos, el daño será exactamente $D_{GAG}$, pero realmente esto es una variable aleatoria, que tendrá sus fluctuaciones ($CV_{D_{GAG}}$).  Si se hacen 2 veces n vuelos de la misma misión, el daño acumulado no será igual en ambos casos, pero la metodología tradicional dice que si, que será $D_{GAG}$, que es en realidad $E[D_{GAG}]$. En cualquier caso, aunque me encaja bastante, no lo tengo del todo maduro (la met.tradicional es un proceso muy complejo y quizás no estoy considerando algo). Si no estamos seguros, podemos dejar simplemente que nuestro ground truth es exactamente $D_{GAG} =  \overline{\mathscr{D}_\text{GAG}}\cdot n$ y no hablar de aproximaciones aquí. Solamente aclarar lo de que la media converge a la media muestral, que es realmente la aproximación que hemos tomado nosotros.}
\item Second, approximating the (true, hidden) expected value of a random variable by its sample mean is a classical estimation method, which holds by virtue of the law of large numbers which assures almost-surely convergence of the sample mean to the true mean for large sample sizes. For illustration, we  assess how the (true, hidden) average per-flight GAG damage deviates from the (estimated) sample average per-flight GAG damage $\overline{\mathscr{D}_\text{GAG}}$ using non-parametric bootstrapping and building the 95\% confidence interval of the bootstrapped mean GAG Damage, as this gives us an uncertainty interval where the true mean will lies with high probability.  In \autoref{fig:bootstrap_means}  we  represent, for a specific PSE, $k_{t}$ and mission ($n \approx 1300$ flights), the distribution of bootstrapped means alongside its 2.5 and 97.5 percentiles as well as the bootstrap average. We also locate the original sample mean for reference.  Observe that the sample mean of the GAG damage, $\overline{\mathscr{D}_\text{GAG}}$  and the average of the bootstrapped means $\mu$ virtually coincide, $\overline{\mathscr{D}_\text{GAG}} \approx \mu = 6.19\times10^{-5}$, and the bootstrap $CI95 = [6.10\times10^{-5},6.29\times10^{-5}]$ is a very narrow one. This validates our procedure  
    %and the standard deviation is small (3.20\% and 0.82\% of the mean respectively) 
    of identifying the true mean with the sample mean. Results hold similarly for other cases.
\end{itemize}

    \begin{figure}[h!]
        \centering
        \includegraphics[width=0.8\linewidth]{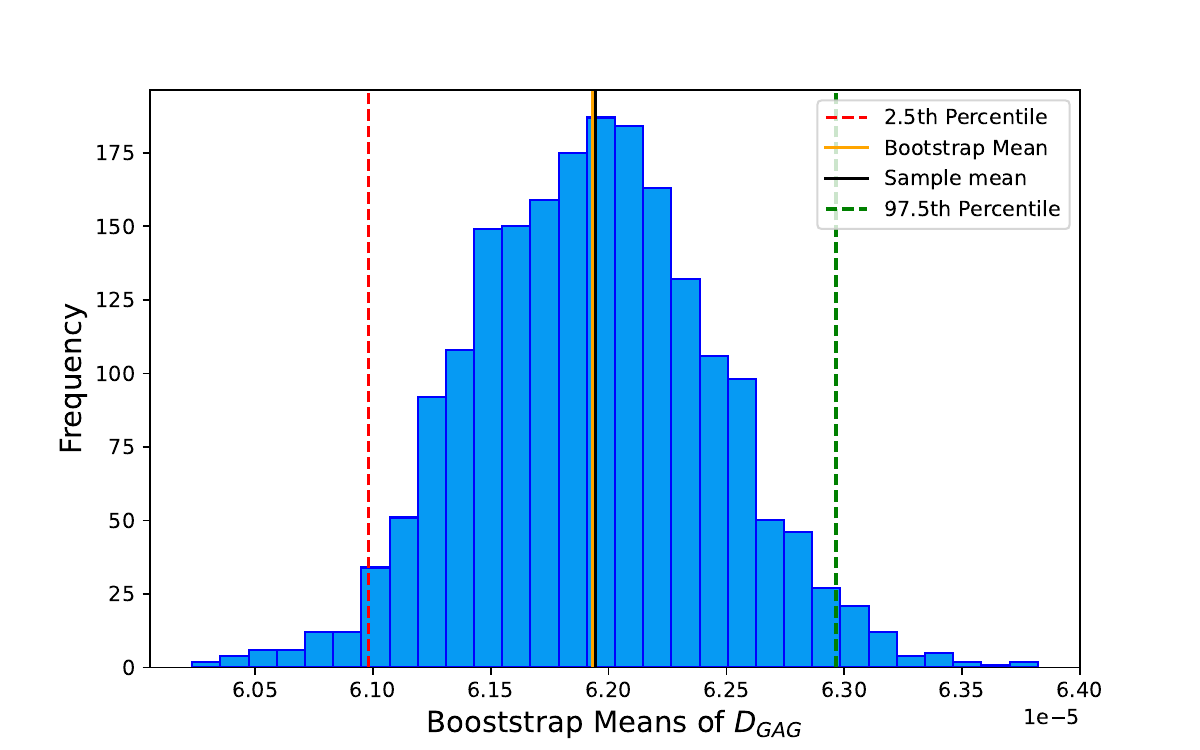}
        \caption{Distribution of bootstrapped means associated to the GAG damage $\mathscr{D}_{\text{GAG}}$, for a specific PSE and mission. The 2.5 and 97.5 percentiles --conforming the bootstrap CI95 of average per-flight GAG damage are highlighted, alongside the average of the bootstrapped means $\mu$ and the sample mean $\overline{\mathscr{D}_{\text{GAG}}}$.}
        \label{fig:bootstrap_means}
    \end{figure}

This procedure critically depends on the number of flights $n$ being moderately large, something that in this work systematically holds as we have $n \in [800, 13000]$. Now, it is always desirable to establish what would be the minimum number of flights $n_\text{min}$  such that our method is acceptable. This is akin to ask for what sample size (i) the first moment closure is no longer a valid one, or (ii) when estimating the true mean with the sample mean is not acceptable. The actual question is difficult to answer in general, as it requires  to define what we mean by `acceptable'. Here we provide two different roadmaps to estimate the minimum $n_{min}$ to meet a certain statistical guarantee, namely, that with a certain significance level $\alpha$ and confidence level $p = 1- \alpha$, the mismatch between the sample mean of the per-flight GAG damage $\overline{\mathscr{D}_\text{GAG}}$ and the true mean is smaller than a certain tolerance $\epsilon$ (measured in relative error units).

In general terms, invoking the central limit theorem (CLT, which holds as high order moments of $\mathscr{D}_\text{GAG}$ do not diverge, see above), such difference drops to zero as fast as $1/\sqrt(N)$. In the event that $\mathscr{D}_\text{GAG}$ is Gaussian, simple probability arguments provide a closed expression for $n_{min}$:

\begin{equation}
    n_{min} = \left(\frac{Z(\alpha/2)\sigma}{\epsilon\overline{\mathscr{D}_\text{GAG}}}\right)^{2},
\end{equation}

where $\alpha$ is our desired significance level, $Z(\alpha/2)$ can be obtained from the values from a standard normal distribution, and sample standard deviation $\sigma$ and the sample mean $\overline{\mathscr{D}_\text{GAG}}$ can be computed from the population dataset. 
When the marginal distribution deviates from a pure Gaussian, the expression above is only an approximation, which nonetheless becomes a tight one as soon as $n$ is moderately large (our case) and if the underlying distribution does not have strong fluctuations around the mean (i.e. long tails), which, again, is the case here. %This implies that the distribution of the means is approximately a normal distribution (confirmed by a Kolmogorov-Smirnov test, KS Statistic: 0.011, P-value: 0.95 for \autoref{fig:bootstrap_means}. )}

For illustration, for the concrete mission and PSE discussed above with $ n\approx 1300 $, and fixing $\epsilon = 0.02 $ and $\alpha = 0.05$  (i.e. allowing for a maximum 2\% relative error with a 95\% confidence), we find $n_{min}$ = 883, which is substantially smaller than the actual number of flights. Similarly,  we can estimate the maximum error associated with the exact $n$ used. In this case, for a 95\% confidence level, the maximum error is $1.64\%$.

\medskip \noindent 

Finally, we can also define an heuristic which makes use of nonparametric bootstrapping (and therefore does not assume any shape for the GAG Damage distribution). The minimal acceptable size of the dataset can be found by (i) establishing an acceptable interval for the sample mean in the original dataset (e.g. the CI95 of the bootstrapped mean), then (ii) sequentially downsampling the original dataset, estimating the new sample mean in each case, and (iii)  detecting when the sample mean is no longer within the established interval. 

In practice, the procedure is implemented as follows: first, we define a new data size $n_{new} < n$. Then, we perform independent resamplings (here we performed 2000, with replacement) of the original dataset of size $n_{new}$, and compute the percentage of cases where the new sample mean of the resampled datasets falls within the originally established interval.\\

These `acceptable intervals' were built with two different approaches, each providing different and complementary information: i) the interval is defined by fixing a priori a relative error around the average of the bootstrapped means $\mu$  (see e.g. \autoref{fig:bootstrap_means}) so as to build the interval $[\mu - \epsilon\mu, \mu + \epsilon\mu]$, and ii) the interval is defined as the 95\% confidence interval of the bootstrapped mean.

\begin{figure}
    \centering
    \includegraphics[width=0.48\linewidth]{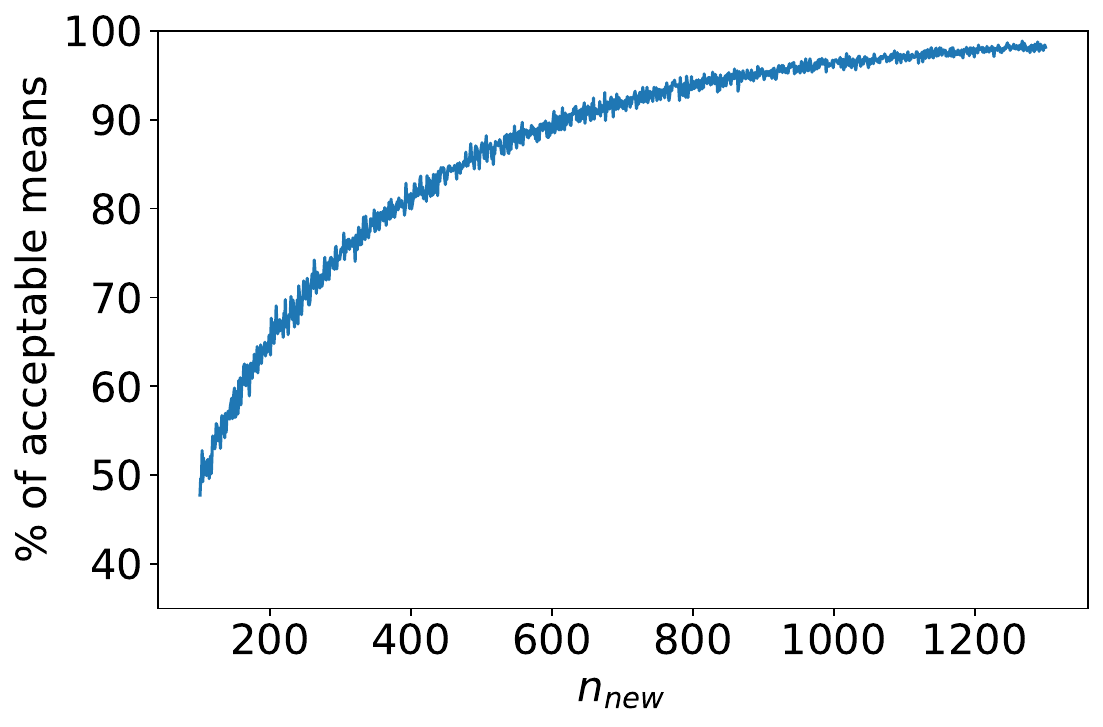}
    \includegraphics[width=0.48\linewidth]{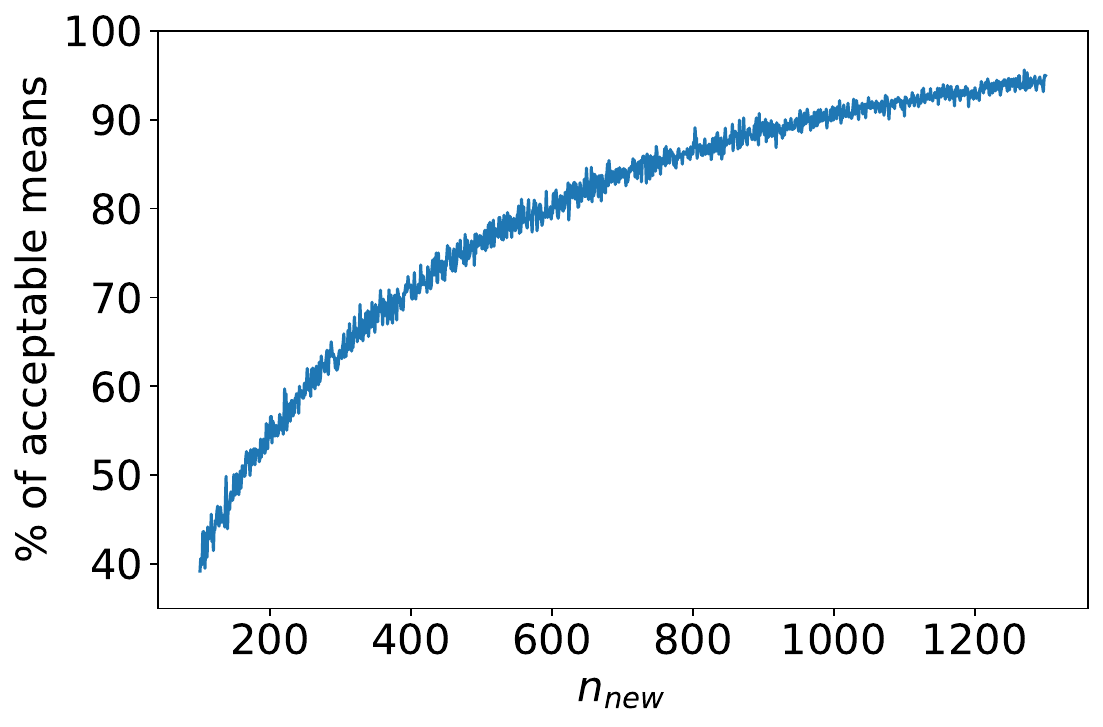}
    \caption{(Left panel) Percentage of resamples with $n_{\text{new}}<n$ data where the sample mean is within the acceptable interval defined initially as $[\mu - \epsilon\mu, \mu +\epsilon\mu]$, as a function of the datasize sample $n_\text{new}$. (Right panel) Similar to the left panel, but here the acceptable interval is defined as the bootstrap mean CI95.}
    \label{fig:n_min}
\end{figure}

For illustration, using the same mission and PSE with $n \approx 1300$, in \autoref{fig:n_min}  we show the percentage of subsets falling within each interval respectively,  as a function of $n_{new}.$ Observe that for a sample size of 883 (flagged as $n_\text{min}$ by the analytical approach above), in about $95\%$ of the resamples the sample average error is equal or lower than the established error, (2\%),  whereas $87.5\%$ of the samples fall in the CI95 interval. %This difference is due to the fact that the percentiles 97.5\% and 2.5\% have a deviation from $\mu$ smaller than 2\%, leading to a smaller interval and a lower percentage of samples falling in it. 

%\begin{figure}
%    \centering
%    \includegraphics[width=0.9\linewidth]{Figures/Results/C95_nnew.pdf}
%    \caption{Minimum sample size vs \% of samples within C95}
%    \label{fig:c95_nnew}
%\end{figure}

%% For citations use: 
%%       \citet{<label>} ==> Lamport (1994)
%%       \citep{<label>} ==> (Lamport, 1994)
%%
% Example citation, See \citet{lamport94}.

%% If you have bib database file and want bibtex to generate the
%% bibitems, please use
%%
%%  \bibliographystyle{elsarticle-harv} 
%%  \bibliography{<your bibdatabase>}

%% else use the following coding to input the bibitems directly in the
%% TeX file.

%% Refer following link for more details about bibliography and citations.
%% https://en.wikibooks.org/wiki/LaTeX/Bibliography_Management

% \begin{thebibliography}{00}

% %% For authoryear reference style
% %% \bibitem[Author(year)]{label}
% %% Text of bibliographic item

% \bibitem[Lamport(1994)]{lamport94}
%   Leslie Lamport,
%   \textit{\LaTeX: a document preparation system},
%   Addison Wesley, Massachusetts,
%   2nd edition,
%   1994.

% \end{thebibliography}
\bibliographystyle{elsarticle-num}
\bibliography{cas-refs}
\end{document}